\documentclass[onefignum,onetabnum]{siamonline171218}
\usepackage{amssymb} %uses R for real numbers
\usepackage{amsmath} %bold symbols in math mode
\usepackage{bbm} % for 1 in conditions
\usepackage{mathtools}
\usepackage{algorithm} %algorithm
\usepackage[noend]{algpseudocode} %algorithm
\usepackage{url} %links in bibliog
\usepackage{array,graphicx}
\usepackage{pgfplots}
\usepackage{booktabs} %for tables
\pgfplotsset{compat=1.9}
\usepgfplotslibrary{units}
\usepackage{tikz} %graphs
\usepackage[numbers,sort]{natbib} %ordering citations
\usepackage{subcaption}
\usepackage{multirow}

\DeclareMathOperator*{\argmin}{argmin}
\DeclareMathOperator*{\argmax}{argmax}

\newcommand{\myvec}[1]{\ensuremath{\begin{pmatrix}#1\end{pmatrix}}}

\theoremstyle{definition}

\newtheorem{prop}{Proposition}[section]
\theoremstyle{plain}

\title{Efficient Relaxations for Dense CRFs with Sparse Higher Order Potentials }
\author{Thomas Joy\thanks{\textit{Indicates equal contribution}} \and Alban Desmaison\footnotemark[1] \and Thalaiyasingam Ajanthan\footnotemark[1] \and Rudy Bunel\and Mathieu Salzmann\and Pushmeet Kohli\and Philip H.S. Torr\and M. Pawan Kumar}

\begin{document}
	
\maketitle

\begin{abstract}
	
	Dense conditional random fields (CRFs) have become a popular framework for modelling several problems in computer vision such as stereo correspondence and multi-class semantic segmentation. By modelling long-range interactions, dense CRFs provide a labelling that captures finer detail than their sparse counterparts. Currently, the state-of-the-art algorithm performs mean-field inference using a filter-based method but fails to provide a strong theoretical guarantee on the quality of the solution.
	
	A question naturally arises as to whether it is possible to obtain a maximum a posteriori (MAP) estimate of a dense CRF using a principled method. Within this paper, we show that this is indeed possible. We will show that, by using a filter-based method, continuous relaxations of the MAP problem can be optimised efficiently using state-of-the-art algorithms. Specifically, we will solve a quadratic programming (QP) relaxation using the Frank-Wolfe algorithm and a linear programming (LP) relaxation by developing a proximal minimisation framework. By exploiting labelling consistency in the higher-order potentials and utilising the filter-based method, we are able to formulate the above algorithms such that each iteration has a complexity linear in the number of classes and random variables. The presented algorithms can be applied to any labelling problem using a dense CRF with sparse higher-order potentials. In this paper, we use semantic segmentation as an example application as it demonstrates the ability of the algorithm to scale to dense CRFs with large dimensions. We perform experiments on the Pascal dataset to indicate that the presented algorithms are able to attain lower energies than the mean-field inference method.
	
\end{abstract}

\section{Introduction}
Conditional random fields (CRFs) are a popular framework for modelling several problems in computer vision.
 The energy function of the CRF consists of a sum of three types of terms: \textit{unary energies} that depend on the label for one random variable; \textit{pairwise energies} that depend on the labels of two random variables; and \textit{higher-order energies} that depend on a collection of random variables. Notable works 
such as~\cite{EfficientContinuousRelaxationsforDenseCRF,EfficientInferenceinFullyConnectedCRFswithGaussianEdgePotentials,EfficientLinearProgrammingforDenseCRFs}
 focus on just the unary and pairwise energies, leaving out the higher-order energies for computational efficiency.

The popularity of CRFs led to a considerable research
effort in efficient energy minimisation algorithms. One of the biggest successes
of this effort was the development of several accurate continuous relaxations of
the underlying discrete optimisation problem~\cite{Ravikumar:2006:QPR:1143844.1143937,Kleinberg:2002:AAC:585265.585268}. An important advantage of
such relaxations is that they lend themselves easily to analysis, which allows
us to compare them theoretically~\cite{Ravikumar:2006:QPR:1143844.1143937}, as well as establish bounds on the
quality of their solutions~\cite{chekuri2001approximation}. However, despite the successes of continuous relaxations, the algorithms used to solve such relaxations fail to scale well with the number of pairwise connections. To combat this aforementioned deficiency, traditional energy minimisation methods employed sparse connectivity structures, such as
4 or 8 connected grid CRFs

By using a mean-field inference method~\cite{kollerbook}, Kr\"{a}henb\"{u}hl and Koltun~\cite{EfficientInferenceinFullyConnectedCRFswithGaussianEdgePotentials} were able to solve CRFs with many pairwise connections. It was shown that the use of dense pairwise connections achieved a more accurate labelling. In order
to operationalise dense CRFs, Kr\"{a}henb\"{u}hl and Koltun~\cite{EfficientInferenceinFullyConnectedCRFswithGaussianEdgePotentials} made two key
observations. First, the pairwise potentials used in computer vision typically
encourage similar labelling. This enabled them to restrict themselves to the
special case of Gaussian pairwise potentials introduced by Tappen et al.~\cite{tappenGMRF}. Second, the message computation required at each iteration
of mean-field can be carried out in $\mathcal{O}(N)$ operations using the filtering
approach of Adams et al. \cite{FastHigh-DimensionalFilteringUsingthePermutohedralLattice}, where $N$ is the number of
random variables (of the order of hundreds of thousands).
 Vineet et al.  \cite{Vineet2014} made use of this filter-based method to perform mean-field inference on a dense CRF
with sparse higher-order potentials, which provided a further improvement in segmentation accuracy. 
%However, the mean-field inference method is an approximation and relies on minimising the KL-divergence between the true distribution and an approximate distribution of the CRF, where the marginals are forced to be independent.

While the mean-field algorithm does not provide any theoretical guarantees on
the quality of the solution, the use of a richer model, namely
dense CRFs with sparse higher-order potentials, still allows us to obtain a significant improvement in the accuracy
of several computer vision applications compared to sparse models~\cite{EfficientInferenceinFullyConnectedCRFswithGaussianEdgePotentials, Vineet2014}.
However, this still leaves open the intriguing possibility that the same
filtering approach - that enabled the efficient mean-field algorithm - can also be
used in conjunction with the principled methods of energy minimisation with continuous relaxations. In this work we show that this is indeed possible, specifically, the main contributions of this paper are as follows:

1$)$ We are the first to combine dense CRFs with higher-order potentials when using a continuous relaxation of the MAP problem. Specifically, we formulate the energy function as both a QP and a LP relaxation and go on to show that both can be optimised efficiently using the filter-based method. As a novel contribution, we then extend the energy minimisation algorithms of our existing work \cite{EfficientContinuousRelaxationsforDenseCRF, EfficientLinearProgrammingforDenseCRFs} to deal with these higher-order potentials, whilst maintaining a complexity that is linear in the number of labels and random variables at each iteration.

2$)$ We provide novel relaxations of the higher-order terms which are based on the $P^n$-Potts model \cite{Kohli07p3&}. These formulations have been tailored to suit the specific relaxation in a way that is amenable to the continuous relaxation of the MAP problem. We also ensure the formulation allows efficient energy minimisation using the two frameworks mentioned above.

In more detail, we make two contributions to the problem of energy
minimisation in dense CRFs with sparse higher-order potentials. First, we show that the conditional gradient of
a quadratic programming (QP) relaxation~\cite{Ravikumar:2006:QPR:1143844.1143937} can be computed in a 
complexity linear in the number of labels and random variables. Together with our observation that the optimal step-size of a descent
direction can be computed analytically, this allows us to minimise the QP
relaxation efficiently using the Frank-Wolfe algorithm~\cite{frankWolfeQuadraticProgramming}. 
Second, we introduce an iterative linear programming (LP) minimization algorithm which has
a complexity at each iteration that is also \text{linear} in the number of labels and random variables.
To this end, instead of relying on a standard
subgradient technique, we propose to make use of the proximal 
method~\cite{parikh2014proximal}. The resulting proximal problem
has a smooth dual, which can be efficiently optimized using block coordinate
descent.
We show that each block of variables can be optimized efficiently.
Specifically, for one block, the problem decomposes into significantly smaller
subproblems, each of which is defined over a single pixel.
For the other block, the problem can be optimized via 
the Frank-Wolfe
algorithm~\cite{frankWolfeQuadraticProgramming,lacoste2012block}. We show that the
conditional gradient required by this algorithm can be computed efficiently.
In particular, we
modify the filtering method of~\cite{FastHigh-DimensionalFilteringUsingthePermutohedralLattice} such that the conditional
gradient can be computed in a complexity \text{linear} in the number of labels and random variables. Besides this linear complexity, our approach has two additional benefits.
% over the subgradient method of~\cite{desmaison2016efficient}. 
First, it can be
initialized with the solution of a faster, less accurate algorithm, such as
mean-field~\cite{EfficientInferenceinFullyConnectedCRFswithGaussianEdgePotentials}, thus speeding up convergence.
Second,
the optimal step size of our iterative procedure can be obtained analytically,
thus overcoming the need to rely on an expensive line search procedure.

There are preliminary versions of this work available, and the interested reader is encouraged to visit ~\cite{EfficientContinuousRelaxationsforDenseCRF, EfficientLinearProgrammingforDenseCRFs}. This work can be considered as a unified view of our previous work \cite{EfficientContinuousRelaxationsforDenseCRF,EfficientLinearProgrammingforDenseCRFs} with a novel addition of higher-order potentials. To keep this paper self-contained all relevant information and findings are detailed in this paper. Specifically, our contribution is a QP and LP relaxation for dense CRF with sparse higher-order potentials and their associated energy minimisation frameworks. 
				
\section{Related Work}
Kr\"{a}henb\"{u}hl and Koltun popularised the use of densely connected CRFs at the pixel level~\cite{EfficientInferenceinFullyConnectedCRFswithGaussianEdgePotentials}, resulting in significant improvements both in terms of the quantitative performance and in terms of the visual quality of their results.
By restricting themselves to Gaussian pairwise potentials, they made the computation of the message passing in mean-field feasible.
This was achieved by formulating message computation as a convolution in a higher-dimensional space, which enabled the use of an efficient filter-based method~\cite{FastHigh-DimensionalFilteringUsingthePermutohedralLattice}. Recent works leverage deep learning to achieve deep embeddings of the pairwise potential \cite{chandra2017dense}, however \cite{chandra2017dense} acts on the patch space whereas in this work we act on the pixel space.

While the initial work by~\cite{EfficientInferenceinFullyConnectedCRFswithGaussianEdgePotentials} used a version of mean-field that is not guaranteed to converge, their follow-up paper~\cite{paramLearning} proposed a convergent mean-field algorithm for negative semi-definite label compatibility functions.
Recently, Baqu{\'{e}} et al.~\cite{princip-parallel} presented a new algorithm that has convergence guarantees in the general case.
Vineet et al.~\cite{Vineet2014} extended the mean-field model to allow the addition of higher-order terms on top of the dense pairwise potentials, enabling the use of co-occurence potentials~\cite{Ladicky:2010:GCB:1888150.1888170} and $P^n$-Potts models~\cite{Kohli07p3&}.

Independently from the mean-field work, Zhang et al.~\cite{Chen:2012:EIF:2354409.2354957} designed a different set of constraints that lends itself to a QP relaxation of the original problem.
Their approach is similar to ours in that they use continuous relaxation to approximate the solution of the original problem but differs in the form of the pairwise potentials.
The algorithm they propose to solve the QP relaxation has linearithmic\footnote{An algorithm is said be linearithmic if it runs in a computational complexity of $\mathcal{O}(Nlog(N))$.} complexity while ours is linear in the number of labels and random variables.
Furthermore, it is not clear whether their approach can be easily generalised to tighter relaxations such as the LP.

Wang et al. \cite{DBLP:journals/corr/WangSH15} derived a semi-definite programming relaxation of the energy minimisation problem, allowing them to reach lower energies than mean-field.
Their approach has the advantage of not being restricted to Gaussian pairwise potentials. Inference is made feasible by performing a low-rank approximation of the Gram matrix of the kernel, instead of using the filter-based method.
However, in theory the complexity of their algorithm is the same as our QP, but in practice the runtime is significantly higher.

The success of the inference algorithms naturally led to interest in methods for learning the parameters of dense CRFs, whilst learning the parameters is orthogonal to this work, we include a brief review for completeness.
Combining them with fully convolutional neural networks~\cite{long2015fully} has resulted in high performance on semantic segmentation applications~\cite{DBLP:journals/corr/ChenPKMY14}.
Several works~\cite{DBLP:journals/corr/SchwingU15,zheng2015conditional} showed independently how to jointly learn the parameters of the unary and pairwise potentials of the CRF.

In this paper, we use the same filter-based method~\cite{FastHigh-DimensionalFilteringUsingthePermutohedralLattice} as the one employed in mean-field.
We use it to solve continuous relaxations of the original problem that have both convergence and quality guarantees.
Our work can be viewed as a complementary direction to previous research trends in dense CRFs.
While~\cite{paramLearning,princip-parallel,Vineet2014} improved mean-field and~\cite{DBLP:journals/corr/SchwingU15,zheng2015conditional} learnt the parameters, we focus on the energy minimisation problem.

\section{Problem Formulation}\label{chap:pf}
	While CRFs can be used for many different applications, we use semantic segmentation as an illustrative example. As will be seen shortly, by using the appropriate choice of random variables, labels and potentials, our model provides an intuitive framework for semantic segmentation.
	
		\subsection{Dense CRF Energy Function}\label{sec:crf}
			We define a dense CRF over a set of $N$ random variables $\mathcal{X} = \{X_1,...,X_N\}$ where each random variable $X_a$ takes a single label from the set of $M$ labels $\mathcal{L} = \{l_1,...,l_M\}$. To formalise this labelling, a vector $\textbf{x} \in \mathcal{L}^N$ is introduced, such that the element $x_a$ of $\textbf{x}$ holds the label associated with the random variable $X_a$. Before proceeding to the energy function, it will prove useful to define a \textit{clique} and its relationship to the sparse higher-order potentials. Formally, a clique is defined as a fully connected subgraph. In the context of this work, a clique with three or more random variables represents a higher-order potential and a clique with two random variables is represented by a pairwise potential. A given clique $S_p$ is a subset of $\mathcal{X}$ and the set of cliques containing higher-order potentials $\mathcal{S}$ is defined below:					 
			\begin{align}
			\label{set_sp} \mathcal{S} & =   \{S_1,...,S_R\}\\
			\text{s.t } S_p & \in  \{S \subseteq \mathcal{X} \mid |S| > 2\}.
			\end{align}
			Here, R represents the total number of cliques in the set $\mathcal{S}$. It will also prove useful to introduce another set $\mathcal{R}_p$, which represents the set of indexes for the random variables in the clique $S_p$, this can be formally expressed as $\mathcal{R}_p = \{a \in \{1,...,N\} \mid X_a \in S_p\}$. With the introduction of $\bold{x}_p$, which is a vector of more than two elements, containing the labels of the random variables in the clique $S_p$, the energy function can be defined as:		
			\begin{align}\label{ip_energy}
			E(\textbf{x}) = \sum_{a=1}^N \phi_{a}(x_a) + \sum_{a=1}^N\sum_{\substack{b=1 \\ b \neq a}}^N\psi_{a,b}(x_a,x_b) + \sum_{p = 1}^R \theta_p(\bold{x}_p),
			\end{align}
			where $\phi_{a}(x_a)$ denotes the unary potential, $\psi_{a,b}(x_a,x_b)$ denotes the pairwise potential, $\theta_p(\bold{x}_p)$ denotes the \text{clique potential}. The {unary potential} represents the cost of assigning the random variable $X_a$ the label $x_a$. The {pairwise potential} represents the cost of assigning the random variables $X_a$ and $X_b$ the labels $x_a$ and $x_b$ respectively. The clique potential represents the cost of assigning all random variables in ${S_p}$ the labels $\mathbf{x}_p$, and embodies the higher-order potentials. 
			
			So far, the dense CRF has been described using random variables and their associated labels. In this work we use semantic segmentation as a tangible application to demonstrate the energy minimisation of the proposed methods. In detail, a random variable corresponds to a pixel and the associated labels correspond to a semantic class. A \textit{superpixel} - which is a collection of homogeneous spatially adjacent pixels - is represented by a higher-order clique. The optimal solution to this energy function forms an optimisation problem over the variable $\bold{x}$ and can be compactly written as:
			\begin{align}
			\bold{x^*} = \argmin_\bold{x \in \mathcal{L}^N} {E(\bold{x})}.
			\end{align}		
			In the general case this minimisation problem is NP-hard \cite{kolmogorov2004energy} and hence cannot be solved in polynomial time. To this extent, efficient methods will be introduced in Section \ref{chap:qp} and \ref{chap:lp} that compute approximate solutions for this minimisation problem. 
					
			\subsubsection{Unary Potentials}\label{sec:potentials}
				 The unary potentials for this formulation can be arbitrary, but generally provide a rough initial labelling solution. In this work we employ unary potentials which are derived from TextonBoost \cite{moreBoost,textonBoost}. More detail on the generation of the unary potentials is given in Section \ref{par:unaries}. %, and for the MSRC data set are freely available online \cite{unary_location}. Kr{\"{a}}henb{\"{u}}hl and Koltun \cite{EfficientInferenceinFullyConnectedCRFswithGaussianEdgePotentials} produced a full set of unary values for the MSRC data set. The unary values were produced via a combination of a 17-dimensional filter bank as proposed by Shoton et al. \cite{textonBoost} and the introduction of colour information, histogram orientated gradients and spatial information following Ladick\'{y} et al. \cite{moreBoost}.					
		    \subsubsection{Gaussian Pairwise Potentials}\label{sec:gaus_pairwise}
					We follow the work of \cite{EfficientInferenceinFullyConnectedCRFswithGaussianEdgePotentials} by using Gaussian Pairwise potentials, taking the form of:
					\begin{align}
					\psi_{a,b}(x_a,x_b) & = \mu(x_a,x_b)\text{K}_{ab}, \\\nonumber
					\text{s.t } \text{K}_{ab} &= \sum_m w^{(m)} k^{(m)}\big(\bold{f}^{(m)}_{a},\bold{f}^{(m)}_{b}\big),
					\end{align}
					where $\mu(x_a,x_b)$ is a scalar representing the \textit{label compatibility}, $\text{K}_{ab}$ is the \textit{pixel compatibility} function which is defined in the next paragraph, $w^{(m)}$ is a scalar weighting factor and $k^{(m)}\big(\bold{f}^{(m)}_{a},\bold{f}^{(m)}_{b}\big)$ are Gaussian kernels taking the form of:
					\begin{align}\label{eq:kernel}
					k^{(m)}(\bold{f}^{(m)}_{a},\bold{f}^{(m)}_{b}) = \text{exp}\bigg(-\frac{||\bold{f}^{(m)}_{a} - \bold{f}^{(m)}_{b} ||^2}{2\sigma_{(m)}^2}\bigg).
					\end{align}
					 The terms $\bold{f}^{(m)}_{a}$ and $\bold{f}^{(m)}_{b}$ are feature vectors containing the spatial and colour information of the image with pixel indices $a$ and $b$ respectively. The value of $\sigma^2_{(m)}$ is the kernel bandwidth. 
					\paragraph{Pixel Compatibility}For multi-class semantic segmentation problems, the pixel compatibility function takes the form of contrast-sensitive two-kernel potentials, defined as:  
					\begin{align}\label{eq:pixel_compatability}
					\text{K}_{ab} = w^{(1)}\exp\Bigg(-\frac{|p_a - p_b|^2}{2\sigma_{(1)}^2} - \frac{|I_a - I_b|^2}{2\sigma_{(2)}^2}\Bigg) + w^{(2)}\exp\Bigg(-\frac{|p_a - p_b|^2}{2\sigma_{(3)}^2}\Bigg),
					\end{align}					
					with $I_a$, $I_b$ and $p_a$, $p_b$ representing the colour information and spatial information of pixels $a$ and $b$ respectively. The first term corresponds to the \textit{bilateral kernel} and is inspired by the observation that pixels of similar colour and position are likely
					to take the same label, the second term corresponds to a \textit{spatial kernel} which penalises small isolated regions. The parameters $w^{(1)}$,$w^{(2)}$,$\sigma_{(1)}^2$,$\sigma_{(2)}^2$ and $\sigma_{(3)}^2$ are obtained via cross-validation, more detail is given in Section \ref{subsec:train}.
					\paragraph{Label Compatibility} The \textit{label compatibility} function $\mu(x_a,x_b)$ forms part of the cost of assigning the random variables $X_a$ and $X_b$ the labels corresponding to the value of $x_a$ and $x_b$ respectively. The {label compatibility} function used for this work is the Potts model \cite{EfficientContinuousRelaxationsforDenseCRF} and is specified as: 				
					\begin{align}\label{eq:potts}
						{\mu}_{Potts}(x_a,x_b) = \mathbbm{1}[x_a\ne x_b],
					\end{align}	
					where $\mathbbm{1}[\cdot]$ is the Iverson bracket. Whilst other label compatibility function exists, such as metric or semi-metric functions~\cite{Kleinberg:2002:AAC:585265.585268}, the Potts model was chosen as it enables more sophisticated minimisation algorithms to be leveraged which will be discussed in Section \ref{chap:qp} and \ref{chap:lp}.
					
			\subsubsection{Higher Order Potentials}\label{subsec:clique}		
			In this work, the higher-order terms are represented as a \textit{clique potential}. We formulate the higher-order potential using the $P^n$-Potts models~\cite{Kohli07p3&}. Specifically, if all of the random variables in $S_p$ do not take the same label, the clique potential introduces a constant cost, which we set to be proportional to the variance of the colour information of the superpixel. The clique potential is defined by:
			\begin{align}\label{eq:clique_energy}
			\theta_p(\bold{x}_p) & =
			\begin{cases}
			0 & \text{if } x_c = x_d \text{,   }  \forall c,d \in \mathcal{R}_p\\
			\Gamma\text{exp}\Big\{\frac{-\sigma_p^2}{\eta}\Big\} & \text{otherwise,}
			\end{cases}
			\end{align}
			where $\Gamma$ and $\eta$ are cross-validated parameters and $\sigma_p^2$ represents the variance of the pixel colour values within the clique $S_p$. To this extent, the set of random variables which form the clique $S_p$  must be carefully chosen. Hence, by context of the image, all of the corresponding pixels in the clique $S_p$ must represent the same object.

			\paragraph{Generating Cliques}
			For this work, a clique represents a super-pixel of an over segmented image. A superpixel is a collection of adjoining pixels who share similar colour information. The cliques were generated using the mean-shift algorithm \cite{Comaniciu02meanshift:} which is a semiparametric method of segmenting an image into superpixels. We used the mean-shift algorithm due to its simplicity, however in practice any algorithm that generates over segmentations can be used. Evaluating the quality of the over segmentation is beyond the scope of this work, however, we cross validate the size of the super-pixels to ensure we use higher-order potentials that match the problem. Representing superpixels by higher-order potentials introduces an implicit constraint on $S_p$, that is $S_p \cap S_q = \varnothing \text{ } \forall p,q \neq p$, as every pixel is assigned to exactly one superpixel. Whilst this is not a necessary constraint (as the algorithms can deal with arbitrary sizes of cliques), we will make use of this later on to ensure that the complexity of each iteration is linear in the number of labels and pixels.
																	
			\subsubsection{Filtering Method}\label{subsec:permuto_lattice}
				The pixel compatibility function defined in (\ref{eq:pixel_compatability}), was chosen to take a Gaussian form due to the fact that it allows a filter-based method~\cite{FastHigh-DimensionalFilteringUsingthePermutohedralLattice} to be utilised. This filter-based method exploits the \textit{permutohedral lattice} to achieve efficient computation of operations featuring Gaussian kernels, specifically it approximates the following:
				\begin{align}\label{eq:permuto_ex}
				\forall a \in \{1,...,N\},\text{\indent}v'_a = \sum_{b=1}^Nk(\bold{f}_a,\bold{f}_b)v_b,
				\end{align}					
				{where $v'_a,v_{b} \in \mathbb{R}$ and $k(\bold{f}_a,\bold{f}_b)$ is a Gaussian kernel described in section \ref{sec:gaus_pairwise}. A na\"ive approach would take  $\mathcal{O}(N^2)$ operations. However, the use of the filtering method enables this operation to be computed in approximately $\mathcal{O}(N)$ operations. Kr{\"{a}}henb{\"{u}}hl and Koltun~\cite{EfficientInferenceinFullyConnectedCRFswithGaussianEdgePotentials} employed this filter-based method to compute the message passing step of the mean-field inference algorithm efficiently. We investigated the accuracy of the filter-based method~\cite{FastHigh-DimensionalFilteringUsingthePermutohedralLattice} with differing values for the variances of equation (\ref{eq:pixel_compatability}) in our preliminary work ~\cite{EfficientContinuousRelaxationsforDenseCRF}. The results indicate that the filtering method introduces an error scaling factor, which for large values of $N$ tends to 0.6. The interested reader is referred to Appendix A of~\cite{EfficientContinuousRelaxationsforDenseCRF} for more information. This scaling factor will be propagated into the gradient, but it is implicitly accounted for when the optimal step size is computed, and hence does not have an adverse affect on the algorithms as the direction of the gradient is exact.
					
			\subsubsection{Integer Program Formulation}\label{subsec:ip_init}
					We now formulate the energy minimisation function (\ref{ip_energy}) as an integer program (IP). To this end, a vector $\bold{y} \in \mathbb{R}^{NM}$ is introduced, such that its elements $y_{a:i} \in \{0,1\}$ are binary variables indicating whether or not the random variable ${X}_a$ takes the label ${l}_i$. The vector $\mathbf{y}_p  = \{y_{c:i} | c \in \mathcal{R}_p\ , i \in \mathcal{L}\}$ is introduced which holds the vectors of indicator variables for $\bold{x}_p$.	With this new notation the energy minimisation function can be defined as:			
					\begin{align}\label{eq:sp_ip}
					\min_\mathbf{y} \sum_{a=1}^N \sum_{i \in \mathcal{L}}\phi_{a}{(i)}{y_{a:i}} + \sum_{a=1}^N\sum_{\substack{b=1 \\ b \neq a}}^N\sum_{i \in \mathcal{L}}\sum_{j \in \mathcal{L}}\psi_{a,b}{(i,j)}y_{a:i}y_{b:j} + \sum_{p=1}^R\overline{\theta}_p(\mathbf{y}_p),\\
					\begin{tabular}{ccccc}\nonumber
					s.t & $\sum_{i \in \mathcal{L}}y_{a:i} = 1$ & $\forall a \in \{1,...,N\}$, & \\
					& $y_{a:i} \in \{0,1\}$ & $\forall a \in \{1,...,N\}$, & $\forall i \in \mathcal{L}$,\\
					\end{tabular}\\
					 \overline{\theta}_p(\mathbf{y}_p) =
					\begin{cases}
					0 & \text{if } y_{c:i} = y_{d:i} \text{,   }  \forall c,d \in \mathcal{R}_p, c\neq d, \forall i \in \mathcal{L}\\
					\Gamma\text{exp}\Big\{\frac{-\sigma_p^2}{\eta}\Big\} & \text{otherwise}.
					\end{cases}
					\end{align}
					The first set of constraints ensure that each random variable has to be assigned exactly one label, whilst the second constraint ensures that the labelling is binary. It is important to note that $\overline{\theta}_p(\cdot)$ is a polynomial with an order equal to number of random variables within the clique $S_p$. Normally the manipulation of $\overline{\theta}_p(\cdot)$ would exhibit an intractable complexity, however by exploiting labelling consistency in the sparse higher-order potentials, it will be shown that this higher-order polynomial can be reformulated in a tractable manner.  
																		
		\subsection{Relaxations}
		It is worth noting that the IP in (\ref{eq:sp_ip}) is NP-hard\cite{kolmogorov2004energy} and hence cannot be solved in polynomial time. We address this issue by relaxing the integral constraint to approximate the IP, enabling us to formulate an energy minimisation problem. Specifically we formulate a QP relaxation and an LP relaxation given in Sections \ref{chap:qp} and \ref{chap:lp} respectively.
		
\section{Quadratic Program}\label{chap:qp}
We are now ready to demonstrate how the filter-based method~\cite{FastHigh-DimensionalFilteringUsingthePermutohedralLattice} can be used to optimise our first continuous relaxation, namely the QP relaxation.
	\subsection{Notation and Formulation}
		The unary and pairwise potentials of the IP given in equation (\ref{eq:sp_ip}) can be neatly summarised in vector form with linear algebra operations. To this extent, the unary potential can be concisely written as the dot product between the vector $\mathbf{y} \in \mathbb{R}^{NM}$ and the vector of unary terms denoted $\boldsymbol{\phi} \in \mathbb{R}^{NM}$. The pairwise potential is a little more complex, and will require the use of the label compatibility matrix $\boldsymbol{\mu}_{Potts} \in \mathbb{R}^{M \times M}$, which in this case is the Potts model described in equation (\ref{eq:potts}). For the pixel compatibility function, each kernel (\ref{eq:kernel}) is represented by the Gram matrix $\boldsymbol{K}^{(m)} \in \mathbb{R}^{N \times N}$. The element of $\boldsymbol{K}^{(m)}$ at index $(a,b)$ corresponds to the value of $k^{(m)}\big(\bold{f}^{(m)}_{a},\bold{f}^{(m)}_{b}\big)$. The matrix $\boldsymbol{\varPsi} \in \mathbb{R}^{NM \times NM}$ represents the pairwise terms and is defined as:
		\begin{align}
	 		\boldsymbol{\varPsi} = \boldsymbol{\mu} _{Potts}\otimes \sum_{m} w^{(m)} \big( \boldsymbol{K}^{(m)} - \boldsymbol{I}_N \big),
		\end{align}		
		where $\otimes$ is the Kronecker product, $\boldsymbol{I}_N$ is the identity matrix of size ${N \times N}$.  Similarly to \cite{paramLearning}, $\boldsymbol{K}^{(m)}$ has a unit diagonal and hence the identity matrix $\boldsymbol{I}_N$ is introduced for completeness. The objective function of the IP for the unary and pairwise potentials is given in vectorized form as:	
		\begin{gather}\label{eq:qp_no_clique}§
		\min_\mathbf{y} \boldsymbol{\phi}^{T}{\mathbf{y}} + {\mathbf{y}^{T}}{\boldsymbol{\varPsi}}{\mathbf{y}},\\
		\begin{tabular}{ccccc}\nonumber
			s.t & $\sum_{i \in \mathcal{L}}y_{a:i} = 1$ & $\forall a \in \{1,...,N\}$, & \\
			& $y_{a:i} \in \{0,1\}$ & $\forall a \in \{1,...,N\}$, & $\forall i \in \mathcal{L}$.\\
		\end{tabular}
		\end{gather}
		In the general case, a clique potential forms a high-order polynomial with an order equal to the number of random variables in each clique. However, by exploiting labelling consistency, we are able to reformulate this high-order polynomial as a lower-order one. To this end, a binary auxiliary variable $z_{p:i}$ is introduced which indicates whether or not all of the random variables in the clique $S_p$ take the label $\mathcal{L}_i$. The auxiliary variable  $z_{p:i}$ is given as:
		\begin{align}
		z_{p:i} =
		\begin{cases}
			0, & \text{if } y_{c:i} = 1,\text{   }\forall c \in \mathcal{R}_p \\
			1, & \text{otherwise.}
		\end{cases}
		\end{align}
		In other words if all random variables in the clique $S_p$ take the same label then $z_{p:i} = 0$.
		Before proceeding to the definition of the clique potential for the QP it will be beneficial to introduce an additional term $H_p(a)$, which is used to indicate if the random variable $X_a$ belongs to the clique $S_p$. Formally $H_p(a) = 1$ if $a \in \mathcal{R}_p$ and $H_p(a) = 0$ otherwise. With the addition of the auxiliary variable $z_{p:i}$ and the indicator term $H_p(a)$, the clique potential forms a quadratic polynomial in $z_{p:i}$ and $y_{a:i}$, which is given below. The clique potential is given as:
		\begin{align}\label{eq:qp_sp_pot}
			f_c := \sum_{p=1}^R\sum_{i \in \mathcal{L}}{C_p}\Big(z_{p:i} +\big((1 - z_{p:i}) \sum_{a=1}^N H_p(a)(1 - y_{a:i})\big)\Big).
		\end{align} 
		It is worth noting that the last term will always evaluate to zero. However, once the binary constraints on $z_{p:i}$ and $y_{a:i}$ are relaxed, the latter term provides a coupling between $z_{p:i}$ and $y_{a:i}$. More detail will be given on this in Section \ref{sec:qp_relax}. The vectorised version of $z_{p:i}$ is $\mathbf{z} \in \mathbb{R}^{MR}$. The values of $H_p(a)$ form the matrix $\boldsymbol{H} \in \mathbb{R}^{MR \times NM}$, which is a sparse matrix of ones, such that the elements are in the correct order to perform the summations. The matrix $\boldsymbol{H}$ is purely provided for illustrative purposes and due to its sparse nature, in the implementation it is not stored as a matrix of size $MR \times NM$. Instead, $R$ arrays are instantiated with each array containing the indexes of the random variables within the clique. With the addition of $\mathbf{z}$ and $\boldsymbol{H}$, the IP  can be concisely written in vector form as:						
		\begin{align}\label{eq:ip}
			\min_\mathbf{y,z}{f(\mathbf{y},\mathbf{z})} &= \min_\mathbf{y,z}\big(\boldsymbol{\phi}^T\mathbf{y} + \mathbf{y}^T\boldsymbol{\varPsi}\mathbf{y} + 
\mathbf{c}^T\mathbf{z} + \mathbf{(1_z - z)}^T\mathbf{C}\boldsymbol{H}\mathbf{(1_y - y)}\big),
		\end{align}
		where $\mathbf{c} \in \mathbb{R}^{MR}$ is a vector containing the constants $C_p$ in the appropriate order. The matrix $\mathbf{C} \in \mathbb{R}^{MR \times MR}$ is the diagonal matrix of the vector $\mathbf{c}$. The vectors $\mathbf{1_z} \in \mathbb{R}^{MR}$ and $\mathbf{1_y} \in  \mathbb{R}^{NM}$ are vectors of all ones.
		\subsubsection{Relaxations}\label{sec:qp_relax}
		The Integer Program introduced in equation (\ref{eq:ip}) is an NP-hard problem. To overcome this difficulty, it is proposed to relax the binary constraints on the indicator variable $y_{a:i}$ and the auxiliary variables $z_{p:i}$, allowing them to take fractional values between 0 and 1. Formally, with these relaxations, the feasible set for $\mathbf{y}$ and $\mathbf{x}$ becomes:
		\begin{align}
		\label{set:set_y_vars}\mathcal{Y} = \left\{
		\bold{y}\text{  }
		\begin{tabular}{|cc}
			$\sum_{i \in \mathcal{L}}y_{a:i} = 1 $&$   a \in \{1,...,N\}$\\
			$y_{a:i}  \geq 0 $&$  a \in \{1,...,N\},  i \in \mathcal{L}$ \\
		\end{tabular}
		\right\},
		\end{align}
		\begin{align}
		\label{set:set_z_vars}\mathcal{Z} = \left\{
		\bold{z}\text{  }
		\begin{tabular}{|cc}
			$0 \leq z_{p:i}  \leq 1 $&$  S_p \in \mathcal{S},  i \in \mathcal{L}$ \\
		\end{tabular}
		\right\}.
		\end{align}
		Thus, the QP relaxation can be formally defined as:
		\begin{align}
		\label{eq:qp_sp}
			\min_\mathbf{y,z}{f(\mathbf{y},\mathbf{z})} &= \min_\mathbf{y,z}\big(\boldsymbol{\phi}^T\mathbf{y} + \mathbf{y}^T\boldsymbol{\varPsi}\mathbf{y} + 
			\mathbf{c}^T\mathbf{z} + \mathbf{(1_z - z)}^T\mathbf{C}\boldsymbol{H}\mathbf{(1_y - y)}\big), 	\\\nonumber
		\text{s.t } \mathbf{y} &\in  \mathcal{ Y}, \mathbf{z} \in  \mathcal{ Z}.
		\end{align}
						
	\subsection{Minimisation} The minimisation of the objective function is achieved via the Frank-Wolfe algorithm \cite{frankWolfeQuadraticProgramming}, which is advantageous for two reasons: Firstly, the Frank-Wolfe algorithm is projection free, and secondly the conditional gradient can be computed in a complexity linear in the number of pixels and labels. The objective function of (\ref{eq:qp_sp}) can be solved in several ways, however, even though (\ref{eq:qp_sp}) is non-convex, we choose to obtain a local minimum using the Frank-Wolfe algorithm, as we are able to take advantage of the aforementioned qualities. Whilst the Frank-Wolfe algorithm normally optimises convex objectives, it has been proven to find a stationary point at a rate of $\mathcal{O}(1/\sqrt{t})$  of a non-convex objective function over a convex compact set\footnote{A topological space $\mathcal{X}$ is called compact if every open cover has a finite subcover. Furthermore, a set $\mathcal{D} \subset \mathbb{R}^N$ is compact if and only if, it is closed and bounded \cite{kelly}.}, where $t$ is the number of iterations \cite{lacoste_nonconvex_FW}. The key steps of the algorithm are shown in Algorithm \ref{algo:fw}. To utilise the Frank-Wolfe algorithm effectively, three steps need to be taken: obtain the gradient of the objective function (step 3); efficient conditional gradient computation (step 4) and the optimal step size calculation (step 5). All three of these requirements are achieved in a feasible manner and details are given in this section.	
	\begin{algorithm}[t]
		\caption{QP Minimisation Algorithm}\label{algo:fw}
		\begin{algorithmic}[1]
			\State $\textbf{y}^{0} \in \mathcal{Y}, \textbf{z}^{0} \in \mathcal{Z}$
			\Comment{Initialise}
			\While {not converged}
			\State	$\textbf{g}^{t} \gets \nabla{f(\textbf{y}^{t},\textbf{z}^{t})}$ \Comment{Compute the gradient}
			\State 	$\myvec{\mathbf{s^{\textit{t}}_y},\mathbf{s^{\textit{t}}_z}}^T \gets \text{argmin}_{\mathbf{s_y} \in \cal{Y}, \mathbf{s_z} \in \cal{Z}}{\big\langle\myvec{\mathbf{s_y},\mathbf{s_z}}^T ,\textbf{g}^{t}\big\rangle}$ \Comment{Compute the conditional gradient}
			\State  $\delta \gets \text{argmin}_{\delta \in [0,1]}{f(\mathbf{y}^{t} + \delta({\mathbf{s^{\textit{t}}_y} - \mathbf{y}^{t})},\mathbf{z}^{t} + \delta{({\mathbf{s^{\textit{t}}_z} - \mathbf{z}^{t})}})}$ \Comment{Compute the optimal step size}
			\State  $\myvec{\mathbf{y}^{t+1},\mathbf{z}^{t+1}}  \gets \myvec{{\textbf{y}^{t} + \delta({\mathbf{s^{\textit{t}}_y} - \mathbf{y}^{t})}}, {\textbf{z}^{t} + \delta({\mathbf{s^{\textit{t}}_z} - \mathbf{z}^{t})}}}$ \Comment{Update}
			\EndWhile
		\end{algorithmic}
	\end{algorithm}
	\subsubsection{Gradient Computation}\label{subsec:grad_comp}
	The Frank-Wolfe algorithm requires efficient computation of the gradient, which can easily be achieved for this problem. Formally, the gradient of $f$ is defined as:
		
	\begin{align}
	\label{eq:qp_div}
	\nabla{f(\textbf{y},\textbf{z})} =
	\myvec{
	\mathbf{\boldsymbol{\phi} + \boldsymbol{2\varPsi}{y} + \boldsymbol{H}^T\mathbf{C}({z} - 1_{\mathbf{z}})}\\
	\mathbf{c} + \mathbf{C}\boldsymbol{H}(\mathbf{y - {1_y}})
	}.
	\end{align} 
	 Specific attention is drawn to the complexity of the gradient in the $\mathbf{y}$ direction. The unary term is left as a constant and hence scales linearly with the number of labels and random variables. Computing the value of the pairwise potential in the na\"{\i}ve way would result in a complexity of the order $\mathcal{O}((MN)^2)$, which for dimensions of an image is intractable. However, due to the elements of $\boldsymbol{\varPsi}$ containing Gaussian kernels, this expensive computation of the pairwise potential can be performed in linear time using the filter-based method, more detail on this filter-based method is given in section \ref{subsec:permuto_lattice}. 
	 
	 Due to the fact that $\boldsymbol{H}$ is implemented as a list of lists data structure and there is no intersection between cliques $S_p \cap S_p = \varnothing$, the resulting complexity of the clique potential is of the order $\mathcal{O}{(NM)}$ as for each clique we perform a sum over only the labels and pixels within that clique.
	 
	 \subsubsection{Low-cost Gradient Computation}\label{sec:free_grad}
	 We observe that the gradient introduced in Section \ref{subsec:grad_comp} need not be explicitly computed at every iteration. Instead the gradient can be incremented from its initial value using the update equations which are given as:
	 \begin{align}
	 \myvec{\textbf{y}^{t+1}\\\textbf{z}^{t+1}} = \myvec{
	 {\textbf{y}^{t} + \delta({\mathbf{s_{y}}^{t} - \textbf{y}^{t})}}\\ 
	 {\textbf{z}^{t} + \delta{(\mathbf{s_{z}}^{t} - \textbf{z}^{t})}}
	 }.
	 \end{align}
	 Where $s_{y}$ and $s_{z}$ are the conditional gradients of $f(\textbf{y},\textbf{z})$. The expensive operations of $2\boldsymbol{\varPsi}\mathbf{y}$ and $\boldsymbol{H}^T\textbf{{C}z}$ in equation (\ref{eq:qp_div}), can be avoided by using the values of $2\boldsymbol{\varPsi}\mathbf{(s_{y} - y)}$ and $\boldsymbol{H}^T\mathbf{C(s_{z} - z)}$ - which are both computed as part of the optimal step size - and by using the update equations. By multiplying the update equation for $\mathbf{y}$ by $2\boldsymbol{\varPsi}$ and multiplying the update equation for $\mathbf{z}$ by $\boldsymbol{H}^T\textbf{C}$, the updated terms can be given as:
	 \begin{align}
	 2\boldsymbol{\varPsi}\mathbf{y}^{t + 1} &=  2\boldsymbol{\varPsi}\textbf{y}^{t} + 2\delta \boldsymbol{\varPsi}({\mathbf{s_{y}}^{t} - \textbf{y}^{t})},\\
	 \boldsymbol{H}^T\mathbf{{C}z}^{t + 1} &= \boldsymbol{H}^T\textbf{{C}z}^{t} + \delta{\boldsymbol{H}^T\mathbf{C}(\mathbf{s_{z}}^{t} - \textbf{z}^{t})}.
	 \end{align}
	 Thus, allowing the explicit computation of the operations of $2\boldsymbol{\varPsi}\mathbf{y}$ and $\boldsymbol{H}^T\textbf{{C}z}$ to be avoided. Instead their values can be incremented from their previous state. Hence the updated gradient in $\mathbf{y}$ is also an increment from the previous step via the addition of $2\delta \boldsymbol{\varPsi}({\mathbf{s_{y}}^{t} - \textbf{y}^{t})} + \delta{\boldsymbol{H}^T\mathbf{C}(\mathbf{s_{z}}^{t} - \textbf{z}^{t})}$ and is more formally given as:
	 \begin{align}
	 \nabla_y{f(\mathbf{y}^{t+1},\mathbf{z}^{t+1})} = \nabla_y{f(\mathbf{y}^t,\mathbf{z}^t)} + 2\delta \boldsymbol{\varPsi}({\mathbf{s_{y}}^{t} - \textbf{y}^{t})} + \delta{\boldsymbol{H}^T\mathbf{C}(\mathbf{s_{z}}^{t} - \textbf{z}^{t})}.
	 \end{align}
	 
	 A similar approach can be taken for $\nabla_z{f(\mathbf{y}^{t+1},\mathbf{z}^{t+1})}$. Incrementing the gradients, reduces the operational complexity by a constant factor of two. This is due to the fact that the filter-based method does not need to be called when computing the gradient and the product of $\boldsymbol{H}^T\textbf{{C}z}$ does not need to be computed either.

	\subsubsection{Conditional Gradient Computation}
	Computing the conditional gradient is an essential step in the Frank-Wolfe algorithm, and we show that it can be computed in a complexity linear in the number of labels and pixels. The conditional gradient $\myvec{\mathbf{s_{y}}\\\mathbf{s_{z}}}, \text{with } \mathbf{s_{y}} \in \mathcal{Y}, \mathbf{s_{z}} \in\mathcal{Z}$, of the objective function $f$ is obtained by solving:
	\begin{align}\label{eq:condgrad_y}
		\myvec{\mathbf{s_{y}}\\\mathbf{s_{z}}} \in \text{argmin}_{\mathbf{s_{y}} \in \mathcal{Y},\mathbf{s_{z}} \in \mathcal{Z}}{\bigg\langle\myvec{\mathbf{s_{y}}\\\mathbf{s_{z}}},\nabla{f(\textbf{y},\textbf{z})}\bigg\rangle}.
	\end{align}
	Minimising equation (\ref{eq:condgrad_y}) with dimensions proportional to that of an image, would normally be an expensive operation. However, the reader's attention is drawn to the fact that the feasible set $\mathcal{Y}$ is linearly separable into $N$ subsets as follows $\mathcal{Y} = \prod_a \mathcal{Y}_a$, where $\mathcal{Y}_a = \{y_{a:i}|\sum_{i \in \mathcal{L}}y_{a:i} = 1, y_{a:i} \geq 0, i \in \mathcal{L} \}$. Exploiting this constraint enables the minimisation problem to be broken down into $N$ smaller minimisation problems for each of the random variables in $\mathcal{X}$. Minimising ${\langle\mathbf{s_{y}},\nabla{f(\textbf{y},\textbf{z})}\rangle}$ with respect to $\mathbf{s_{y}}$ is thus achieved via $N$ linear searches with the search space restricted to the number of labels. The resulting computational complexity  of the conditional gradient is $\mathcal{O}(NM)$ and is more formally defined as:	
	\begin{align}
		s^{(y)}_{a:i} =
		\begin{cases}
		1 & \text{if } i = \text{argmin}_{i\in\mathcal{L}}\frac{\partial{f(\mathbf{y},\mathbf{z})}}{\partial{y_{a:i}}}\\
		0 & \text{otherwise}.
		\end{cases}
	\end{align}	
	For the case where $\text{argmin}_{i\in\mathcal{L}}\frac{\partial{f(\mathbf{y},\mathbf{z})}}{\partial{y_{a:i}}}$ yields multiple values, we arbitrarily assign $s^{(y)}_{a:i} = 1 $ for only one of the given minima. The feasible set $\mathcal{Z}$ is also separable and can be decomposed as follows: $\mathcal{Z} = \prod_{p,i} \mathcal{Z}_{p:i}$. Thus the minimisation for $\mathbf{s_{z}} = \text{argmin}_{i\in\mathcal{Z}}\langle\mathbf{s_{z}},\nabla_z{f(\textbf{y},\textbf{z})}\rangle$ can be performed via a linear search through all $MR$ elements. With the constraints on the set $\mathcal{Z}_{p:i} = \{ z | 0 \leq z_{p:i}  \leq 1\ p \in \{1,...,R\}, i \in \mathcal{L}\}$, the conditional gradient $\mathbf{s_{z}}$, is given as:
		\begin{align}
		s^{(z)}_{p:i} = 
		\begin{cases}
		1 & \text{if } \nabla_z{f(\textbf{y},\textbf{z})} < 0\\
		0 & \text{otherwise}.
		\end{cases}
		\end{align}
	The complexity of $\mathbf{s_{z}}$ will always be significantly less than the complexity of $\mathbf{s_{y}}$ due to $R \ll N$. Hence, the computational complexity of calculating the conditional gradient is $\mathcal{O}(NM)$ as computing $\mathbf{s_{y}}$ requires the most floating point operations. 
	\subsubsection{Optimal Step Size Calculation}\label{subsec:opt_step}
	Traditionally, the step size to the Frank-Wolfe algorithm is achieved via line search. However, for this problem the optimal step size can be computed via minimising a quadratic function over a single variable. This quadratic function has a closed form solution and the minimum can be calculated analytically. The optimal step size for the Frank-Wolfe algorithm is obtained by solving:
	\begin{align}
		\delta = \text{argmin}_{\delta\in[0,1]}{f(\mathbf{y} + \delta(\mathbf{s_{y}} - \mathbf{y}),\mathbf{z} + \delta(\mathbf{s_{z}} - \mathbf{z}))},
	\end{align}
	A closed form solution of the optimal step size is given in Appendix \ref{app:step_size}. 
	Obtaining the optimal step size will result in faster convergence and hence yield an efficient algorithm. 
	
	\subsection{Summary}
	The above procedure remains linear in the number of pixels and labels at each iteration, despite introducing higher-order potentials which would normally cause intractability within the algorithm. This is achieved via exploiting the filter-based method \cite{FastHigh-DimensionalFilteringUsingthePermutohedralLattice}, labelling consistency within a clique and enforcing the intersection between cliques to be an empty set. It is worth noting that the filter-based method is called only once per iteration, resulting in an efficient QP minimisation algorithm. 
  	
  \section{Linear Program}\label{chap:lp}
 		  In this section we introduce the LP relaxation, our second continuous relaxation. To this end, relaxations will be applied to the objective function (\ref{eq:sp_ip}) and dual variables will be introduced, allowing the Lagrange dual problem to be formulated. An optimal solution can then be found via the use of the proximal minimisation algorithm \cite{parikh2014proximal} which guarantees a monotonic decrease in the objective function. 
		  	
		  \subsection{Linear Programming Relaxation}
		  In a similar manner to the QP, we also relax the binary indicator variables $y_{a:i}$, and due to the use of the Potts model and the $P^n$-Potts model, we can write the relaxation of (\ref{eq:sp_ip}) as a piecewise linear function, defined as:
		  \begin{align}\label{eq:piece_lp}
		  \min_\mathbf{y} \tilde{E}(\mathbf{y}) = \sum_{a=1}^N \sum_{i \in\cal{L}}\phi_{a:i}{y_{a:i}} &+ \sum_{a=1}^N\sum_{\substack{b=1 \\ b \neq a}}^N\sum_{i \in \mathcal{L}}\text{K}_{ab}\frac{|y_{a:i} - y_{b:i}|}{2} + \sum_{p=1}^RC_p\max_{i\in\mathcal{L}}\max_{\substack{c,d \in \mathcal{R}_p \\ c\neq d }}|y_{c:i} - y_{d:i}|,\\\nonumber
		  \text{s.t  } \mathbf{y} &\in \mathcal{Y},
		  \end{align}   
		  where $\text{K}_{ab}$ is the pixel compatibility function defined in equation (\ref{eq:pixel_compatability}). 
		  For integer labellings, the objective $\tilde{E}(\bold{y})$ has the same value as
		  the IP objective $E(\bold{y})$ and is known to provide the best theoretical bounds \cite{Kleinberg:2002:AAC:585265.585268}. Using standard solvers to minimize this LP would require the introduction of 
		  $\mathcal{O}((NM)^2)$ variables (see equation (\ref{eq:lp_primal})), making it
		  intractable. Therefore the non-smooth objective of equation (\ref{eq:piece_lp}) has to be optimized directly. This
		  was handled using projected subgradient descent
		  in our previous version \cite{EfficientContinuousRelaxationsforDenseCRF}, which also turns out to be inefficient in
		  practice. In this paper, we extend the
		  algorithm introduced in \cite{EfficientLinearProgrammingforDenseCRFs} to handle higher-order potentials, whilst maintaining \text{linear} scaling at each iteration in
		  both space and time complexity.  
		  
		  The piecewise linear functions $|y_{a:i} - y_{b:i}|$, in the pairwise and clique potentials, can be reformulated as piecewise maximum functions $\max\{y_{a:i} - y_{b:i},y_{b:i} - y_{a:i}\}$, and then subsequently replaced by auxiliary variables $v_{ab:i}$ and $w_p$ in the standard way. The auxiliary variables and their constraints enable the minimisation problem to be defined without the piecewise maximum operators. With the introduction of these auxiliary variables the primal minimisation problem can be written as an LP-relaxation and is given as:
		  \begin{align}\label{eq:lp_primal}
		  \min_\mathbf{y,v,w} \sum_{a=1}^N \sum_{i \in\cal{L}}&\phi_{a:i}{y_{a:i}} + \sum_{a=1}^N\sum_{\substack{b=1 \\ b \neq a}}^N\sum_{i \in \mathcal{L}}\frac{\text{K}_{ab}}{2}v_{ab:i} + \sum_{p=1}^RC_pw_p + \frac{1}{2\lambda}||\mathbf{y} - \mathbf{y}^k||^2,\\\nonumber
		  \text{s.t  } v_{ab:i} &\geq y_{a:i} - y_{b:i} \indent\forall a,b \in \{1,...,N\} \indent a \neq b \indent\forall i \in \mathcal{L},  \\\nonumber
		  v_{ab:i} & \geq y_{b:i} - y_{a:i}\indent\forall a,b \in \{1,...,N\} \indent a \neq b \indent\forall i \in \mathcal{L},  \\\nonumber
		  w_p & \geq y_{c:pi} - y_{d:pi}\quad\forall c,d \in \mathcal{R}_p \indent c \neq d \indent\forall i \in \mathcal{L}\indent\forall p \in \{1,...,R\},  \\\nonumber
		  w_p & \geq y_{d:pi} - y_{c:pi}\quad\forall c,d \in \mathcal{R}_p \indent c \neq d \indent\forall i \in \mathcal{L}\indent\forall p \in \{1,...,R\}, \\\nonumber
		  y_{a:i}  & \geq 0\text{\space}\indent\indent\indent\forall a \in \{1,...,N\} \indent\forall i \in \mathcal{L},\\\nonumber
		  \sum_{i \in \mathcal{L}}y_{a:i}  &= 1\text{\space}\indent\indent\indent\forall a \in \{1,...,N\}.\nonumber
		  \end{align}
		  In the next section we present our minimisation strategy for the above LP-relaxation.
		 
		  \subsection{Minimisation}
		  In this section we present our efficient minimisation strategy, which uses the proximal method \cite{parikh2014proximal}. The complexity of each iteration of our implementation remains linear in the number of labels and pixels.
		  \subsubsection{Proximal Minimisation for LP Relaxation}
		  
		  Our goal is to design an efficient minimization strategy for the LP relaxation
		  in~\eqref{eq:piece_lp}. In our previous version \cite{EfficientContinuousRelaxationsforDenseCRF}, we utilised projected subgradient descent to minimise a similar LP to equation~\eqref{eq:lp_primal}, however this method resulted in a significantly high runtime, and a complexity that scales at $\mathcal{O}(MN\text{log}(N))$ at each iteration. To this end, we propose to use the proximal minimization
		  algorithm~\cite{parikh2014proximal}.  The additional quadratic regularization
		  term makes the dual problem smooth, enabling the use of more sophisticated
		  optimization methods. Furthermore, this method guarantees a monotonic decrease in the
		  objective value, enabling us to leverage faster methods for
		  initialization. In the remainder of this paper, we detail
		  this approach and show that each iteration has a complexity linear in the number of labels and pixels. In
		  practice, our algorithm converges in a small number of iterations, thereby
		  making the overall approach computationally efficient. The proximal minimization algorithm~\cite{parikh2014proximal} is an iterative
		  method that, given the current estimate of the solution $\mathbf{y}^k$, solves the
		  problem 
		  \begin{align}\label{eq:lp_prox}
		  \min_\mathbf{y} \tilde{E}(\mathbf{y}) &+ \frac{1}{2\lambda}||\mathbf{y} - \mathbf{y}^k||^2,\\\nonumber
		  \text{s.t } \mathbf{y} &\in \mathcal{Y},
		  \end{align}
		  where $\lambda$ influences the weighting of the quadratic regulariser. In this section we introduce a new algorithm that is tailored to this problem. In particular, we solve the Lagrange dual of (\ref{eq:lp_prox}) in a block-wise fashion.
		  \subsubsection{Dual Formulation}
		 To write the LP relaxation (\ref{eq:piece_lp}) as the dual function four vectors of dual variables will be introduced for each constraint. Namely,
		 \begin{align}
		 \label{eq:alpha}\boldsymbol{\alpha} &= \{\alpha_{ab:i}^1,\alpha_{ab:i}^2 |\text{ } {a} \in \{1,...,N\}\text{, } {b} \in \{1,...,N\}\text{, } a \neq b \text{, }i \in \mathcal{L}\},\\
		 \label{eq:mu}\boldsymbol{\mu} &= \{\mu_{cd:pi}^1,\mu_{cd:pi}^2 |\text{ }  c,d \in \mathcal{R}_p, c \neq d, i \in \mathcal{L}, p \in \{1,...,R\}\},\\
		 \label{eq:gamma}\boldsymbol{\gamma} &= \{\gamma_a |\text{ } a \in \{1,...,N\}, i \in \mathcal{L}\}\},\\
		 \label{eq:bet}\boldsymbol{\beta} &= \{\beta_a |\text{ } a \in \{1,...,N\}\}.
		 \end{align} 
		  Where equation (\ref{eq:alpha}) is for the constraints on $v_{ab:i}$; equation (\ref{eq:mu}) for the constraints on $w_p$; equation (\ref{eq:gamma}) for the non-negativity of $y_{a:i}$ and equation (\ref{eq:bet}) for the labelling of $y_{a:i}$ respectively. The dimensions of these vectors are: $\boldsymbol{\alpha} \in \mathbb{R}^{2N(N-1)M}, \boldsymbol{\mu} \in \mathbb{R}^{2N(N-1)M}, \boldsymbol{\beta} \in \mathbb{R}^{N}$ and $\boldsymbol{\gamma} \in \mathbb{R}^{NM}$. Clearly when dealing with images, the dimensions of $\boldsymbol{\alpha}$ and $\boldsymbol{\mu}$ are intractable. It will be shown that these vectors need not be stored explicitly, instead they can be stored in a compact form. To this extent, three matrices are introduced: $\mathbf{A} \in \mathbb{R}^{NM \times 2N(N-1)M}$,  $\mathbf{U} \in \mathbb{R}^{NM \times 2N(N-1)M}$ and $\mathbf{B} \in \mathbb{R}^{NM \times N}$, such that: 
		  \begin{align}\label{eq:deff_A}
		  (\mathbf{A}\boldsymbol{\alpha})_{a:i} &= \sum_{\substack{b = 1 \\ a \neq b}}^N(\alpha_{ab:i}^2 - \alpha_{ab:i}^1 - \alpha_{ba:i}^2 + \alpha_{ba:i}^1)\\
		  \label{eq:deff_U}(\mathbf{U}\boldsymbol{\mu})_{c:pi} &= \sum_{\substack{d \in \mathcal{R}_p \\ c \neq d}}(\mu_{cd:pi}^2 - \mu_{cd:pi}^1 - \mu_{dc:pi}^2 + \mu_{dc:pi}^1)\\
		  \label{eq:deff_B}(\mathbf{B}\boldsymbol{\beta})_{a:i} &= \beta_a,
		  \end{align}
		  As will be seen shortly, only the products of $(\mathbf{A}\boldsymbol{\alpha}) \in \mathbb{R}^{NM}$, $(\mathbf{U}\boldsymbol{\mu}) \in \mathbb{R}^{NM}$ need to be stored, enabling an efficient implementation. It is also worth defining two of the properties of the matrix $\mathbf{B}$, the product of $\mathbf{B}^T\mathbf{y} = \mathbf{1}$, where $\mathbf{y} \in \mathcal{ Y}$ and $\mathbf{1}$ is a vector of all ones. The second property of $\mathbf{B}$ is that $\mathbf{B}^T\mathbf{B} = M\mathbf{I}$, where $\mathbf{I}$ is the identity matrix and $M$ is the number of labels. With the dual variables introduced it is now possible to proceed to the formation of the dual problem of equation (\ref{eq:lp_primal}).
			\clearpage
		  \begin{prop}\normalfont{Formation of the Lagrange Dual}
		  	\begin{enumerate}
		  		\setlength{\itemindent}{.5in}
		  	\item\textit{The Lagrange dual of equation (\ref{eq:lp_primal}) is given as:}
		  \begin{align}\label{eq:lp_dual}
		  \min_{\boldsymbol{\alpha},\boldsymbol{\mu},\boldsymbol{\beta},\boldsymbol{\gamma}} g(\boldsymbol{\alpha},\boldsymbol{\mu},\boldsymbol{\beta},\boldsymbol{\gamma}) =& \frac{\lambda}{2}|| \mathbf{A}\boldsymbol{\alpha} + \mathbf{U}\boldsymbol{\mu} + \mathbf{B}\boldsymbol{\beta} + \boldsymbol{\gamma} - \boldsymbol{\phi}||^2 + \langle\mathbf{A}\boldsymbol{\alpha} + \mathbf{U}\boldsymbol{\mu} + \mathbf{B}\boldsymbol{\beta} + \boldsymbol{\gamma} - \boldsymbol{\phi}, \mathbf{y}^k \rangle \\\nonumber - &\langle\mathbf{1},\boldsymbol{\beta}\rangle\\
		  \text{s.t }\indent \gamma_{a:i}  \geq& \text{  } 0  \text{  } \forall a \in \{1,...,N\} \text{  } \forall i \in \mathcal{L}\nonumber,\\	
		  \boldsymbol{\alpha} \in \mathcal{A} =& \left\{
		  \boldsymbol{\alpha}\text{  }
		  \begin{tabular}{|cccc}\nonumber
		  $\alpha_{ab:i}^1 + \alpha_{ab:i}^2 = \frac{K_{ab}}{2}$ & $ a,b \in \{1,...,N\}, a \neq b,  i \in \mathcal{L}$ \\
		  $\alpha_{ab:i}^1,\alpha_{ab:i}^2 \geq 0$ & $  a,b \in \{1,...,N\}, a \neq b,  i \in \mathcal{L}$ \\
		  \end{tabular}
		  \right\},\\
		  \boldsymbol{\mu} \in \mathcal{U} =& \left\{
		  \boldsymbol{\mu}\text{  }\nonumber
		  \begin{tabular}{|c}
		  $\sum_{i \in \mathcal{L}}\sum_{\substack{c,d \in \mathcal{R}_p \\ c \neq d}}\mu_{cd:pi}^1 + \mu_{cd:pi}^2 = C_{p}\quad p \in \{1,...,R\}$\\
		  $\mu_{cd:pi}^1, \mu_{cd:pi}^2 \geq 0$  \quad   $ c,d \in \mathcal{R}_p, c \neq d, i \in \mathcal{L}, p \in \{1,...,R\} $\\
		  \end{tabular}
		  \right\}.\\ \nonumber
		  \end{align}
		  \item \textit{The primal variable $\mathbf{y}$ satisfies the following:}
		  \begin{align}\label{eq:y_feas}
		  \mathbf{y} = \lambda(\mathbf{A}\boldsymbol{\alpha} + \mathbf{U}\boldsymbol{\mu} + \mathbf{B}\boldsymbol{\beta} + \boldsymbol{\gamma} - \boldsymbol{\phi}) + \mathbf{y}^k
		  \end{align}
		  \end{enumerate}
	      \end{prop}
		  \textit{Proof.} A detailed formulation of the Lagrangian and the dual is given in Appendix \ref{app:lp_dual}.
		  \subsubsection{LP Minimisation Algorithm}  
		  The dual problem~(\ref{eq:lp_dual}), in its standard form, can only
		  be tackled using projected gradient descent. However, by separating the
		  variables based on the type of the feasible domains, we are able to formulate an efficient
		  block coordinate descent approach. Each of these blocks are amenable to more
		  sophisticated optimization methods, resulting in a computationally efficient algorithm. 
		  As the dual problem is strictly convex and smooth, the optimal solution is still
		  guaranteed. The variables are separated as follows: $\boldsymbol{\alpha}$ and $\boldsymbol{\mu}$ 
		  into one block and $\boldsymbol{\gamma}$ and $\boldsymbol{\beta}$ into another block, with 
		  each block being amenable to more sophisticated optimisation algorithms. For $\boldsymbol{\beta}$ and 
		  $\boldsymbol{\gamma}$ the problem decomposes over the random variables. Then with the optimal values of $\boldsymbol{\beta}$ and $\boldsymbol{\gamma}$, the minimisation of $\boldsymbol{\alpha}$ and $\boldsymbol{\mu}$ is over a compact domain, and can be efficiently tackled using  the Frank-Wolfe algorithm \cite{frankWolfeQuadraticProgramming}. The complete algorithm is summarised in Algorithm \ref{algo:lp}. 
		  \begin{algorithm}[t]
		  	\caption{Proximal minimisation of LP}\label{algo:lp}
		  	\begin{algorithmic}[1]
		  		\State$\textbf{y}^{0} \in \mathcal{Y}$
		  		\Comment{Initialise}
		  		\For{$k\gets0  ... K$}
		  		\State$\mathbf{A}\boldsymbol{\alpha}^0 \gets \mathbf{0}, \mathbf{U}\boldsymbol{\mu}^0 \gets \mathbf{0}, \mathbf{B}\boldsymbol{\beta}^0 \gets \mathbf{0},  \boldsymbol{\gamma}^0 \gets \mathbf{0}$
		  		\Comment{Initialise}
		  		\For{$t\gets0...T$}
		  		\State $(\boldsymbol{\beta}^t,\boldsymbol{\gamma}^t) \gets \argmin_{\boldsymbol{\beta},\boldsymbol{\gamma}} g(\boldsymbol{\alpha}^t,\boldsymbol{\mu}^t,\boldsymbol{\beta},\boldsymbol{\gamma})$ \Comment{Optimise $\boldsymbol{\beta}^t$ and  $\boldsymbol{\gamma}^t$}
		  		\State $\mathbf{\tilde{y}}^t = \lambda(\mathbf{A}\boldsymbol{\alpha}^t + \mathbf{U}\boldsymbol{\mu}^t + \mathbf{B}\boldsymbol{\beta}^t + \boldsymbol{\gamma}^t - \boldsymbol{\phi}) + \mathbf{y}^k$ \Comment{Update feasible solution}
		  		\State $\myvec{\mathbf{s}_\alpha^t, \mathbf{s}_\mu^t} \gets \argmin_{\mathbf{s}_\alpha \in\mathcal{A}, \mathbf{s}_\mu \in\mathcal{U}}\big\langle\myvec{\mathbf{s}_\alpha, \mathbf{s}_\mu},\nabla g(\boldsymbol{\alpha}^t,\boldsymbol{\mu}^t,\boldsymbol{\beta}^t,\boldsymbol{\gamma}^t)\big\rangle$ \Comment{Conditional gradient}
		  		\State $\delta \gets \argmin_\delta g(\boldsymbol{\alpha}^t + \delta(\mathbf{s}^t_\alpha - \boldsymbol{\alpha}^t),\boldsymbol{\mu}^t + \delta(\mathbf{s}^t_\mu - \boldsymbol{\mu}^t),\boldsymbol{\beta}^t,\boldsymbol{\gamma}^t)$ \Comment{Optimal step size}
		  		\State $\myvec{\boldsymbol{\alpha}^{t+1} , \boldsymbol{\mu}^{t+1}} \gets \myvec{\boldsymbol{\alpha}^{t} + \delta(\mathbf{s}^t_\alpha - \boldsymbol{\alpha}^t),  \boldsymbol{\mu}^{t} + \delta(\mathbf{s}^t_\mu - \boldsymbol{\mu}^t)}$ \Comment{Update}
		  		\EndFor
		  		\State $\mathbf{y}^{k+1} \gets P_\mathcal{Y}({\mathbf{\tilde{y}}^t})$  \Comment{Project the primal solution onto the feasible set $\mathcal{Y}$}
		  		\EndFor
		  	\end{algorithmic}
		  \end{algorithm}
		  \paragraph{Optimising over $\boldsymbol{\beta}$ and $\boldsymbol{\gamma}$}
		  The values of $\boldsymbol{\beta}$ and $\boldsymbol{\gamma}$ are efficiently optimised in linear time with the variables $\boldsymbol{\alpha}$ and $\boldsymbol{\mu}$ fixed as $\boldsymbol{\alpha}^t$ and $\boldsymbol{\mu}^t$. This is achieved via the use of simultaneous equations and the QP minimisation algorithm detailed in \cite{xiao2014multiplicative}.  Due to the unconstrained nature of $\boldsymbol{\beta}$, the minimum value of the dual objective $g$ is obtained when $\nabla_\beta g(\boldsymbol{\alpha^t},\boldsymbol{\mu^t},\boldsymbol{\beta},\boldsymbol{\gamma}) = 0$ and hence $\boldsymbol{\beta}$ can be derived as a function of $\boldsymbol{\gamma}$. 
		  \begin{prop}\normalfont{Optimal for $\boldsymbol{\beta}$ }
		  	\begin{enumerate}
		  		\setlength{\itemindent}{.5in}\item\textit{The optimal value for $\boldsymbol{\beta}$ forms a compact expression given as:}
		  \begin{align}\label{eq:beta}
		  \boldsymbol{\beta} = - \frac{\mathbf{B}^T}{M}(\mathbf{A}\boldsymbol{\alpha^t} + \mathbf{U}\boldsymbol{\mu^t} + \boldsymbol{\gamma} - \boldsymbol{\phi}),
		  \end{align}
		\end{enumerate}
		\end{prop}
			\textit{Proof.} A detailed formulation of the optimal expression for $\boldsymbol{\beta}$ is given in Appendix \ref{app:beta_op}.
			
		   By substituting the expression for $\boldsymbol{\beta}$ into the dual objective (\ref{eq:lp_dual}), a quadratic optimisation problem over $\boldsymbol{\gamma}$ is formed. Interestingly, the resulting problem can be optimized independently for each pixel, with each subproblem being an $M$ dimensional quadratic program (QP) with nonnegativity constraints.
		   \begin{prop}\normalfont{Optimising $\boldsymbol{\gamma}$}
		   	\begin{enumerate}
		   		\setlength{\itemindent}{.5in}\item\textit{The optimal value for $\boldsymbol{\gamma}_a$ is obtained by minimising the following QP:}
		   		\begin{align}\label{eq:lp_qp}
		   		\min_{ \gamma_a} \indent&\frac{1}{2}\boldsymbol{\gamma^T}_a\mathbf{Q}\boldsymbol{\gamma}_a + \langle \mathbf{Q}((\mathbf{A}\boldsymbol{\alpha^t}_a) + (\mathbf{U}\boldsymbol{\mu^t})_a - \boldsymbol{\phi}_a) + \mathbf{y}^k,\boldsymbol{\gamma}_a\rangle,\\
		   		\text{s.t }\indent & \boldsymbol{\gamma}_a \geq \mathbf{0}.\nonumber
		   		\end{align} 
		   	\end{enumerate}
		   \end{prop}
	   	\textit{Proof.} The derivation of equation (\ref{eq:lp_qp}) is given in Appendix \ref{app:gamma_op}.
	   	
		  Here, $\boldsymbol{\gamma}_a$ denotes the vector $\{\gamma_{a:i}|i \in \mathcal{L}\}$ and $\mathbf{Q} = \lambda(\boldsymbol{I} - \frac{\mathbf{1}}{M}) \in \mathbb{R}^{M \times M}$.  For notational simplicity it will be beneficial to write the quadratic program above (\ref{eq:lp_qp}) in the following way:
		   
		  \begin{align}\label{eq:lp_qp_h}
		  \min_{\gamma_a} \indent&\frac{1}{2}\boldsymbol{\gamma^T}_a\mathbf{Q}\boldsymbol{\gamma}_a - \langle \mathbf{h}_a,\boldsymbol{\gamma}_a\rangle,\\
		  \mathbf{h}_a =  - &\mathbf{Q}((\mathbf{A}\boldsymbol{\alpha^t}_a) + (\mathbf{U}\boldsymbol{\mu^t})_a - \boldsymbol{\phi}_a) - \mathbf{y^k}.\nonumber
		  \end{align} 
		  We optimise each of these sub-problems using the iterative method given in \cite{xiao2014multiplicative}, as it enables the optimisation to remain linear in the number of labels at each iteration. The key stage of the algorithm is the element-wise update equation, which is given by:
		  \begin{align}
		  \gamma_{a:i} = \gamma_{a:i}\Bigg[\frac{2(\mathbf{Q}^-\gamma_a)_i + h^+_{a:i} + c}{(|\mathbf{Q}|\gamma_a)_i + h^-_{a:i} + c}\Bigg],
		  \end{align}
		  where $\mathbf{Q}^- = \max(-\mathbf{Q}, \mathbf{0})$, $|\mathbf{Q}| = \text{abs}(\mathbf{Q})$, $h^+_{a:i} = \max(h_{a:i},0)$, $h^-_{a:i} = \max(-h_{a:i},0)$ and $0 < c \ll 1$. Once an optimal value for $\boldsymbol{\gamma}$ is obtained, the value of $\boldsymbol{\beta}$ can be calculated via equation (\ref{eq:beta}). Note that, even though the matrix $\mathbf{Q}$ has $M^2$ elements, the multiplication by $\mathbf{Q}$ can be performed in $\mathcal{O}(M)$. In
		  particular, the multiplication by $\mathbf{Q}$ can be decoupled to a multiplication by an
		  identity matrix and a matrix of all ones, both of which can be performed in
		  linear time. Similar observations can be made for the matrices $\mathbf{Q}^-$ and $|\mathbf{Q}|$,
		  hence the time complexity of the above update is $\mathcal{O}(M)$. The interested reader is referred to \cite{xiao2014multiplicative} for more information. 
		  
		  Once a value for $\boldsymbol{\gamma}$ and $\boldsymbol{\beta}$ are obtained, the values for $\boldsymbol{\gamma}^t$ and $\boldsymbol{\beta}^t$ are fixed as $\boldsymbol{\gamma}$ and $\boldsymbol{\beta}$. Due to the fact that optimisation of $\boldsymbol{\gamma}$ decomposes over the number a pixels, and the optimisation of each subproblem is linear in the number of lables, the total complexity of the optimisation of $\boldsymbol{\gamma}$ and $\boldsymbol{\beta}$ is linear in the number of labels and random variables at each iteration.
		  
		  \paragraph{Optimising over $\boldsymbol{\alpha}$ and $\boldsymbol{\mu}$}
		  We now turn to the problem of optimizing over $\boldsymbol{\alpha}$ and $\boldsymbol{\mu}$ given $\boldsymbol{\beta}^t$ and $\boldsymbol{\gamma}^t$.
		  To this end, we use the Frank-Wolfe algorithm~\cite{frankWolfeQuadraticProgramming}, which has the advantage of being projection free. Furthermore, we show that the conditional gradient can be computed in a linear complexity and that the step size can be obtained analytically. The time-complexity of each iteration of this method is linear in the number of pixels and labels. In practice, the Frank-Wolfe algorithm is only run for a fixed number of iterations.
		 
		  \paragraph{Conditional Gradient Computation}
		  With the dual variables fixed at $\boldsymbol{\alpha}^t,\boldsymbol{\mu}^t,\boldsymbol{\beta}^t,\boldsymbol{\gamma}^t$ the conditional gradient $\myvec{\mathbf{s}_\alpha\\\mathbf{s}_\mu}$ is obtained by solving the following:
		  \begin{align}
		  \myvec{\mathbf{s}_\alpha\\\mathbf{s}_\mu} \in \argmin_{\mathbf{s}_\alpha \in\mathcal{A},\mathbf{s}_\mu \in\mathcal{U}}\bigg\langle\myvec{\mathbf{s}_\alpha\\\mathbf{s}_\mu},\myvec{\nabla_{\boldsymbol{\alpha}}g(\boldsymbol{\alpha}^t,\boldsymbol{\mu}^t,\boldsymbol{\beta}^t,\boldsymbol{\gamma}^t) \\ 
		  	\nabla_{\boldsymbol{\mu}}g(\boldsymbol{\alpha}^t,\boldsymbol{\mu}^t,\boldsymbol{\beta}^t,\boldsymbol{\gamma}^t)}\bigg\rangle.
		  \end{align} 
		  Minimising this equation to obtain the conditional gradients $\mathbf{s}_\alpha$ and $\mathbf{s}_\mu$ can be neatly summarised by exploiting the properties of the matrices $\mathbf{A}$ and $\mathbf{U}$ given in equations (\ref{eq:deff_A}) and (\ref{eq:deff_U}) respectively. 
		  \begin{prop}\normalfont \text{Conditional gradient computation}
		  	\begin{enumerate}
		  		\setlength{\itemindent}{.5in}
		  	\item\textit{{The conditional gradient} $\mathbf{s}_\alpha$ {is given by}:}
		  \begin{align}\label{eq:ca_condgrad}
		  (\mathbf{As_\alpha})_{a:i} = \sum_{b=1}^N(K_{ab}\mathbbm{1}[\tilde{y}^t_{a:i} \leq \tilde{y}^t_{b:i}] - K_{ab}\mathbbm{1}[\tilde{y}^t_{b:i} \leq \tilde{y}^t_{a:i}]),
		  \end{align}
		  \item\textit{{The conditional gradient} $\mathbf{s}_\mu$ {is given by}:}
		  %\begin{align}
		  %(\mathbf{Uc})_{c:pi} = \sum_{d}(K_p\mathbbm{1}[\tilde{y}^t_{c:pi} \leq \tilde{y}^t_{d:pi}] - K_p\mathbbm{1}[\tilde{y}^t_{d:pi} \leq \tilde{y}^t_{c:pi}]),
		  %\end{align}
		  \begin{align}
		  (\mathbf{Us_\mu})_{c:pi} = \begin{cases}
		  C_p & \text{if  } \tilde{y}^t_{c:pi} \leq \tilde{y}^t_{d:pj} \textit{ } \forall d \in \mathcal{R}_p\backslash c \textit{, } \forall j \in \mathcal{L}\\
		  -C_p & \text{if  } \tilde{y}^t_{c:pi} \geq \tilde{y}^t_{d:pj} \textit{ } \forall d \in \mathcal{R}_p\backslash c \textit{, } \forall j \in \mathcal{L}\\
		  0 & \text{otherwise},
		  \end{cases}
		  \end{align}
      	  \end{enumerate}
		  {where $\mathbf{\tilde{y}^t}$ is the current (infeasible) solution computed using equation (\ref{eq:y_feas}).}\\
		  \textit{Proof.} Full derivations of the conditional gradients are given in Appendix \ref{app:grad_comp_lp}.
		  \end{prop}
	  		
		  \indent Note that the conditional gradient in (\ref{eq:ca_condgrad}) takes the same form as the subgradient in equation (20) of~\cite{EfficientContinuousRelaxationsforDenseCRF}. This is not a surprising result, as there has been a proven duality relationship between subgradients and conditional gradients for certain problems~\cite{duality}. The conditional gradient $\mathbf{s}_\alpha$ is obtained via the use of a modified version of the advanced filter-based method with more detail given in Appendix \ref{app:filter}, which reduces equation (\ref{eq:ca_condgrad}) to a linear complexity. The conditional gradient $\mathbf{s}_\mu$  is obtained via a linear search through all the elements of each clique to find the minimum and the maximum values for $\tilde{y}^t_{c:i}$ in each clique and setting the values to $C_p$ and $-C_p$ respectively. Hence, the resulting complexity of the conditional gradient is linear in the number of variables and labels. 
		  
		  \paragraph{Optimal Step Size}
		  The performance of any gradient descent based algorithm is fundamentally dependant on the choice of step size. Here, the optimal step size can be computed via minimising a quadratic function over a single variable, which has a closed form solution. This further improves the efficiency of this method. The optimal step size for the Frank-Wolfe algorithm is obtained by solving:

		  \begin{align}
		  \delta = \argmin_{\delta \in [0,1]} g(\boldsymbol{\alpha}^t + \delta(\mathbf{s}^t_\alpha - \boldsymbol{\alpha}^t),\boldsymbol{\mu}^t + \delta(\mathbf{s}^t_\mu - \boldsymbol{\mu}^t),\boldsymbol{\beta}^t,\boldsymbol{\gamma}^t),
		  \end{align}
		  which can be obtained analytically and has a closed form solution.
		  \begin{prop}\normalfont \text{Optimal step size calculation}	
		  		\setlength{\itemindent}{.5in}
		  		\item \textit{The optimal step size to the Frank-Wolfe algorithm is given as:}
		  \begin{align}
		  \delta = P_{[0,1]}\Bigg[\frac{\langle\mathbf{A}\boldsymbol{\alpha}^t + \mathbf{U}\boldsymbol{\mu}^t  - \mathbf{As}^t_\mu -  \mathbf{Us}^t _\alpha ,\mathbf{{y}^t}\rangle}{\lambda||\mathbf{A}\boldsymbol{\alpha}^t + \mathbf{U}\boldsymbol{\mu}^t  - \mathbf{As}^t_\mu -  \mathbf{Us}^t _\alpha||^2}\Bigg],
		  \end{align}
		  \end{prop}
	  		\textit{Proof.} Full derivations of the conditional gradients are given in Appendix \ref{app:lp_step}.
	  		
		  Where $P_{[0,1]}$ indicates the truncation of the quotient into the interval $[0,1]$. 
		  
		 \subsection{Summary}
		 
		 To summarise, our method has the following desirable qualities of an efficient
		 iterative algorithm. With our choice of a quadratic proximal
		 term, the dual of the proximal problem can be efficiently optimized in a block-wise
		 fashion. Specifically, the dual variables $\boldsymbol{\beta}$ and $\boldsymbol{\gamma}$ are computed
		 efficiently by minimising a small QP (of dimensions equal the
		 number of labels) for each pixel independently.
		 The remaining dual variables $\boldsymbol{\alpha}$ and $\boldsymbol{\mu}$ are optimized using the
		 Frank-Wolfe algorithm, where the conditional gradients are computed in linear
		 time, and the optimal step size is obtained analytically. Overall, the time
		 complexity of one iteration of our algorithm is $\mathcal{O}(NM)$ and has no dependence on the number of cliques or their size. This is achieved via again exploiting the filter-based method \cite{FastHigh-DimensionalFilteringUsingthePermutohedralLattice}, labelling consistency within a clique and enforcing the intersection between cliques to be an empty set. 
		 To the best of our knowledge, this constitutes the first LP minimization
		 algorithm for dense CRFs with sparse higher-order potentials, with a complexity linear in the number of labels and pixels per iteration.

\section{Evaluation}

	 This section details the evaluation of the QP and LP implementation outlined in the previous sections, specifically we provide details on: datasets, methods and results. We use semantic segmentation as an example application to demonstrate the low energies achieved by the optimisation methods, and their ability to tackle dense CRFs whose dimensions match those of images. Whilst we are aware that current state-of-the-art methods focus on increasing the Intersection over Union score \cite{zheng2015conditional, chen2018deeplab}, we consider this avenue to be tangential to our work and focus primarily on energy minimisation to evaluate the methods. The experiments were conducted on the Pascal VOC 2010 dataset \cite{pascal-voc-2010} and MSRC dataset \cite{MSRC-dataset}, which are both standard benchmarks for semantic segmentation. 
	 
	 Pascal contains 1928 colour images with dimensions of approximately 500 $\times$ 400 pixels, and 20 classes excluding the background. We split the data in the same way as in \cite{EfficientInferenceinFullyConnectedCRFswithGaussianEdgePotentials}, which is 40\% for training, 15\% for validation and 45\% for testing. We also use the unary potentials from \cite{EfficientInferenceinFullyConnectedCRFswithGaussianEdgePotentials} which were trained using the 40\% training set. To obtain the kernel and higher-order potential parameters, we use the 15\% validation set, with the evaluation performed on the 45\% testing set.
	 
	 MSRC contains 591 images with 21 classes, the dimensions of the images are  320 $\times$ 213 pixels. The labelling ground truths provided in the MSRC data set are of poor quality as regions around the object are left unlabelled and the boundaries are inaccurate. Hence, the current data set is not sufficient for performance evaluation, to overcome this Kr{\"{a}}henb{\"{u}}hl and Koltun \cite{EfficientInferenceinFullyConnectedCRFswithGaussianEdgePotentials} manually produced accurate segmentations for 94 images. It is this smaller dataset with accurate ground truths on which we perform the cross validation and tests. We also use the unary potentials from \cite{EfficientInferenceinFullyConnectedCRFswithGaussianEdgePotentials} which were not trained on any of the images with accurate ground truths.

	 	 We denote the QP and LP implementations as \textbf{QP\textsubscript{clique}} and \textbf{LP\textsubscript{clique}} respectively. We also performed experiments for the QP and LP without introducing higher-order potentials, \textit{i.e} the objective function just consists of a unary and a pairwise potential. Which we denote as \textbf{QP} and \textbf{LP} respectively. To provide a standard benchmark, we compare our methods against methods that optimise a dense CRF model, namely the mean-field algorithm \cite{EfficientInferenceinFullyConnectedCRFswithGaussianEdgePotentials} and it's higher-order variant \cite{Vineet2014}. We denote these methods as \textbf{MF5} and \textbf{MF5\textsubscript{clique}}, which were both run for five iterations as is often done in practice. All experiments were conducted on a 3.60GHz Intel i7-6850K processor. No GPU parallelisation was utilised and the experiments were performed within a single processing thread. The initial starting points for the algorithms are obtained by minimizing the unary potentials.

%	 \subsection{MSRC Dataset}
%	 The experiments were conducted on the MSRC \cite{MSRC-dataset} data set which is a standard benchmark for semantic segmentation. The data set contains 591 images with 21 classes, the dimensions of the images are  320 $\times$ 213 pixels. The labelling ground truths provided in the MSRC data set are of poor quality as regions around the object are left unlabelled and the boundaries are inaccurate. Hence, the current data set is not sufficient for performance evaluation, to overcome this Kr{\"{a}}henb{\"{u}}hl and Koltun \cite{EfficientInferenceinFullyConnectedCRFswithGaussianEdgePotentials} manually produced accurate segmentations for 94 images. It is this smaller dataset with accurate ground truths on which we perform the cross validation and tests. The size of the dataset is of little significance due to the fact that the application of this work is energy minimisation of dense CRFs with sparse higher-order potentials, hence there is little advantage of using datasets which are applicable to deep learning such as PASCAL VOC \cite{pascal-voc-2012} and MS-COCO \cite{microsoft-coco-common-objects-in-context}. Thus the primary metric used for evaluation is the comparison of the resulting average energies.

	 \subsection{Methods}\label{subsec:train}
	 \paragraph{Training of Unary Potentials}\label{par:unaries}
	 The unary potentials were trained using the JointBoost algorithm \cite{jointboost} by Kr{\"{a}}henb{\"{u}}hl and Koltun \cite{EfficientInferenceinFullyConnectedCRFswithGaussianEdgePotentials}. To train the unary potentials for both datasets, $45\%$ of each of the original datasets were used. 
	 
	 \paragraph{Generating Cliques}
	 To generate the higher-order potentials we used the mean-shift algorithm \cite{Comaniciu02meanshift:}. To obtain higher-order potentials that match the problem, we cross validate the minimum region size. We set a spatial and range resolution to 8 and 4 respectively to avoid cross validating a large number of parameters.
	 
	 \paragraph{Cross Validation of Parameters} We use cross validation to find a set of parameters that best represent the semantic segmentation problem. We consider the choice of optimal parameters to be beyond the scope of this work as we are primarily focussing on energy minimisation. Due to the long runtime of the LP, we only performed cross validation on the QP and mean-field algorithms. For the \textbf{QP\textsubscript{clique}} and \textbf{MF5\textsubscript{clique}}, eight parameters had to be cross validated - five for the pixel compatibility function (\ref{eq:pixel_compatability}) and then three for the clique potential (\ref{eq:clique_energy}). For the \textbf{QP}} and \textbf{MF5}, only five parameters had to be cross validated for the pixel compatibility function (\ref{eq:pixel_compatability}). This was achieved using the Spearmint package \cite{Snoek12practicalbayesian}, which uses Bayesian inference to obtain a set of suitable parameters. The cross-validated parameters for the Pascal and MSRC datasets are given in Appendix \ref{app:params}.

	 \subsection{Results} 
	 The collected results provide a quantitative measure of: accuracy, energy and IoU. Accuracy is measured as a percentage of correctly labelled pixels. Energy is the value of the energy function for the resultant labelling. For \textbf{QP}, \textbf{LP} and \textbf{MF5} the assignment energy is calculated using only the unary and pairwise terms of (\ref{eq:sp_ip}), whilst \textbf{QP\textsubscript{clique}}, \textbf{LP\textsubscript{clique}} and \textbf{MF5\textsubscript{clique}} take the energy function of (\ref{eq:sp_ip}). The IoU (Intersection over Union) gives a representation of the proportion of correctly labelled pixels to all pixels taking that class. The optimisation process relies on relaxing the constraints on the variables, allowing them to take fractional values. To manifest the fractional solution as an integral solution the, \textit{argmax} rounding scheme is used, specifically $x_a = \argmax_i(y_{a:i})$. 
	 
\begin{figure}
	 	\begin{tabular}{cc}
	 	\begin{tikzpicture}[scale=0.85]
	 	\centering %image 4_3_s was used
	 	\begin{axis}[
	 	title = {Dense CRF},
	 	xlabel = {Time (s)},
	 	ylabel = {Energy},
	 	xmin = {0},
	 	xmax = {60},
	 	legend pos = north east,]
	 	]	
	 	\addplot[
	 	color=green,
	 	mark=cirlce,
	 	]
	 	coordinates {
	 		(0,6.53083e+07)(0.186197,5.77976e+07)(0.363628,5.57127e+07)(0.56594,5.40496e+07)(0.784529,5.24797e+07)(10,5.24797e+07)(20,5.24797e+07)(30,5.24797e+07)(40,5.24797e+07)(50,5.24797e+07)(60,5.24797e+07)
	 	};
	 	\addlegendentry{MF5}
	 	\addplot[
	 	color=red,
	 	mark=cirlce,
	 	]
	 	coordinates {
	 		(0,8.28222e+07)(0,8.28222e+07)(0,6.42631e+07)(0.218504,6.10876e+07)(0.428072,5.86782e+07)(0.637507,5.68643e+07)(0.847225,5.52864e+07)(1.05675,5.36344e+07)(1.26643,5.20172e+07)(1.47554,5.06966e+07)(1.68539,4.945e+07)(1.89916,4.80669e+07)(2.10893,4.67233e+07)(2.31947,4.52721e+07)(2.52903,4.39144e+07)(2.73866,4.26351e+07)(2.95117,4.14521e+07)(3.16172,4.03178e+07)(3.37178,3.94643e+07)(3.58128,3.8602e+07)(3.79109,3.78516e+07)(4.0013,3.72006e+07)(4.21061,3.67905e+07)(4.41903,3.64884e+07)(4.63093,3.614e+07)(4.83931,3.58264e+07)(5.04978,3.55976e+07)(5.26191,3.53227e+07)(5.47751,3.50933e+07)(5.68704,3.48844e+07)(5.89662,3.46826e+07)(6.10562,3.45e+07)(6.31517,3.42842e+07)(6.52493,3.4092e+07)(6.73369,3.38979e+07)(6.96089,3.36613e+07)(7.17007,3.34524e+07)(7.37842,3.32679e+07)(7.58692,3.31223e+07)(7.79569,3.29401e+07)(8.00424,3.27878e+07)(8.21465,3.26559e+07)(8.43132,3.24894e+07)(8.64344,3.23429e+07)(8.85195,3.22346e+07)(9.06081,3.21621e+07)(9.26984,3.21186e+07)(9.47927,3.20926e+07)(9.69606,3.20697e+07)(9.90872,3.2062e+07)(10.1228,3.2062e+07)(10.3326,3.2062e+07)(10.5409,3.20512e+07)(10.7506,3.20334e+07)(10.9636,3.20249e+07)(11.1755,3.20249e+07)(11.397,3.20249e+07)(11.621,3.20249e+07)(11.8442,3.20249e+07)(12.0675,3.20249e+07)(12.2903,3.20249e+07)(12.5143,3.20249e+07)(12.7378,3.20249e+07)(12.9651,3.20249e+07)(13.1883,3.20249e+07)(13.4164,3.20249e+07)(13.6422,3.20249e+07)(13.8682,3.20249e+07)(14.0907,3.20249e+07)(14.3153,3.20249e+07)(14.5393,3.20249e+07)(14.7633,3.20249e+07)(14.9866,3.20249e+07)(15.2105,3.20249e+07)(15.4351,3.20249e+07)(15.6586,3.20249e+07)(15.8802,3.20249e+07)(16.1037,3.20249e+07)(16.3268,3.20249e+07)(16.5504,3.20249e+07)(16.7749,3.20249e+07)(16.9979,3.20249e+07)(17.2201,3.20249e+07)(17.4432,3.20249e+07)(17.6635,3.20249e+07)(17.8858,3.20249e+07)(18.11,3.20249e+07)(18.3332,3.20249e+07)(18.5564,3.20249e+07)(18.7809,3.20249e+07)(19.0056,3.20249e+07)(19.2329,3.20249e+07)(19.4561,3.20249e+07)(19.6788,3.20249e+07)(19.9024,3.20249e+07)(20.1253,3.20249e+07)(20.3502,3.20249e+07)(20.5745,3.20249e+07)(20.7981,3.20249e+07)(21.0202,3.20249e+07)(21.2442,3.20249e+07)(21.4679,3.20249e+07)(21.6918,3.20249e+07)(21.9142,3.20249e+07)(22.1378,3.20249e+07)(22.3669,3.20249e+07)(22.5912,3.20249e+07)(22.8144,3.20249e+07)(23.0374,3.20249e+07)(23.2612,3.20249e+07)(23.4856,3.20249e+07)(23.7101,3.20249e+07)(23.9338,3.20249e+07)(24.157,3.20249e+07)(24.3795,3.20249e+07)(24.6035,3.20249e+07)(24.8277,3.20249e+07)(25.052,3.20249e+07)(25.2754,3.20249e+07)(25.4997,3.20249e+07)(25.7235,3.20249e+07)(25.9466,3.20249e+07)(26.1698,3.20249e+07)(26.3936,3.20249e+07)(26.617,3.20249e+07)(26.8414,3.20249e+07)(27.0669,3.20249e+07)(27.2912,3.20249e+07)(27.5146,3.20249e+07)(27.7386,3.20249e+07)(27.965,3.20249e+07)(28.2033,3.20249e+07)(28.4288,3.20249e+07)(28.6525,3.20249e+07)(28.876,3.20249e+07)(29.1003,3.20249e+07)(29.3251,3.20249e+07)(29.5482,3.20249e+07)(29.7729,3.20249e+07)(30.0163,3.20249e+07)(30.24,3.20249e+07)(30.4635,3.20249e+07)(30.6873,3.20249e+07)(30.91,3.20249e+07)(31.1333,3.20249e+07)(31.3612,3.20249e+07)(31.5837,3.20249e+07)(31.8073,3.20249e+07)(32.0308,3.20249e+07)(32.2552,3.20249e+07)(32.4782,3.20249e+07)(32.7022,3.20249e+07)(32.9252,3.20249e+07)(33.1494,3.20249e+07)(33.3725,3.20249e+07)(33.5972,3.20249e+07)(33.8202,3.20249e+07)(34.0432,3.20249e+07)(34.2702,3.20249e+07)(34.4927,3.20249e+07)(34.7164,3.20249e+07)(34.9395,3.20249e+07)(35.1624,3.20249e+07)(35.386,3.20249e+07)(35.6086,3.20249e+07)(35.8314,3.20249e+07)(36.0544,3.20249e+07)(36.2788,3.20249e+07)(36.502,3.20249e+07)(36.7257,3.20249e+07)(36.9485,3.20249e+07)(37.1768,3.20249e+07)(37.3998,3.20249e+07)(37.6228,3.20249e+07)(37.8465,3.20249e+07)(38.0695,3.20249e+07)(38.2932,3.20249e+07)(38.5164,3.20249e+07)(38.7386,3.20249e+07)(38.9611,3.20249e+07)(39.1839,3.20249e+07)(39.4074,3.20249e+07)(39.6309,3.20249e+07)(39.8529,3.20249e+07)(40.0821,3.20249e+07)(40.3058,3.20249e+07)(40.5273,3.20249e+07)(40.75,3.20249e+07)(40.9735,3.20249e+07)(41.1966,3.20249e+07)(41.4201,3.20249e+07)(41.6414,3.20249e+07)(41.8653,3.20249e+07)(42.0885,3.20249e+07)(42.3123,3.20249e+07)(42.5354,3.20249e+07)(42.7595,3.20249e+07)(42.9872,3.20249e+07)(43.2116,3.20249e+07)(43.4353,3.20249e+07)(43.6584,3.20249e+07)(43.882,3.20249e+07)(44.1057,3.20249e+07)(44.3294,3.20249e+07)(44.5544,3.20249e+07)(44.7771,3.20249e+07)(45.001,3.20249e+07)(45.2241,3.20249e+07)(45.4478,3.20249e+07)(45.6726,3.20249e+07)(45.9021,3.20249e+07)(46.1249,3.20249e+07)(46.3485,3.20249e+07)(46.5716,3.20249e+07)(46.7948,3.20249e+07)(47.0177,3.20249e+07)(47.2412,3.20249e+07)(47.4643,3.20249e+07)(47.6879,3.20249e+07)(47.9126,3.20249e+07)(48.1359,3.20249e+07)(48.3604,3.20249e+07)(48.5841,3.20249e+07)(48.8137,3.20249e+07)(49.0374,3.20249e+07)(49.2614,3.20249e+07)(49.4845,3.20249e+07)(49.7078,3.20249e+07)(49.9308,3.20249e+07)(50.1565,3.20249e+07)(50.3799,3.20249e+07)(50.6068,3.20249e+07)(50.8299,3.20249e+07)(51.0528,3.20249e+07)(51.2764,3.20249e+07)(51.5002,3.20249e+07)(51.7276,3.20249e+07)(51.9509,3.20249e+07)(52.1737,3.20249e+07)(52.3967,3.20249e+07)(52.6198,3.20249e+07)(52.849,3.20249e+07)(53.0715,3.20249e+07)(53.2953,3.20249e+07)(53.5185,3.20249e+07)(53.7414,3.20249e+07)(53.9652,3.20249e+07)(54.1886,3.20249e+07)(54.4123,3.20249e+07)(54.6412,3.20249e+07)(54.8656,3.20249e+07)(55.0925,3.20249e+07)(55.3155,3.20249e+07)(55.5397,3.20249e+07)(55.7633,3.20249e+07)(55.9905,3.20249e+07)(56.216,3.20249e+07)(56.4421,3.20249e+07)(56.6664,3.20249e+07)(56.8912,3.20249e+07)(57.1155,3.20249e+07)(57.356,3.20249e+07)(57.585,3.20249e+07)(57.8091,3.20249e+07)(58.0407,3.20249e+07)(58.2639,3.20249e+07)(58.4908,3.20249e+07)(58.7172,3.20249e+07)(58.9411,3.20249e+07)(59.1652,3.20249e+07)(59.4008,3.20249e+07)(59.626,3.20249e+07)(59.8549,3.20249e+07)
	 		
	 	};
	 	\addlegendentry{QP}
	 	\addplot[
	 	color=blue,
	 	mark=cirlce,
	 	]
	 	coordinates {
	 		(0,8.28222e+07)(22.2819,1.76533e+07)(44.858,1.4724e+07)(67.4832,1.36901e+07)
	 		
	 	};
	 	\addlegendentry{LP}
	 	
	 	\end{axis}
	 	\end{tikzpicture}
	 	\begin{tikzpicture}[scale=0.85]
	 	\centering %image 8_19_s was used
	 	\begin{axis}[
	 	title = {Dense CRF with high-order terms},
	 	xlabel = {Time (s)},
	 	ylabel = {Energy},
	 	xmin = {0},
	 	xmax = {50},
	 	legend pos = north east,]
	 	]	
	 	\addplot[
	 	color=green,
	 	mark=cirlce,
	 	]
	 	coordinates {
	 		(0,7.84055e+07)(0.262247,6.94966e+07)(0.507834,6.6147e+07)(0.780832,6.32497e+07)(1.04737,6.02577e+07)(10,6.02577e+07)(20,6.02577e+07)(30,6.02577e+07)(40,6.02577e+07)(50,6.02577e+07)(60,6.02577e+07)
	 		
	 	};
	 	\addlegendentry{MF5\textsubscript{clique}}
	 	\addplot[
	 	color=red,
	 	mark=cirlce,
	 	]
	 	coordinates {
	 		(0,7.81273e+07)(0.254786,7.36081e+07)(0.501713,6.99783e+07)(0.749708,6.67588e+07)(0.998445,6.37233e+07)(1.24612,6.10156e+07)(1.49338,5.83603e+07)(1.74122,5.56629e+07)(1.98782,5.31109e+07)(2.23252,5.06141e+07)(2.47934,4.79691e+07)(2.72793,4.56681e+07)(2.97616,4.37518e+07)(3.22387,4.23335e+07)(3.47027,4.10874e+07)(3.73644,4.02363e+07)(3.9831,3.94185e+07)(4.23136,3.86378e+07)(4.47938,3.79447e+07)(4.74209,3.74111e+07)(4.98865,3.69346e+07)(5.23563,3.67205e+07)(5.48261,3.65684e+07)(5.72927,3.63932e+07)(5.97932,3.62284e+07)(6.2282,3.61007e+07)(6.4768,3.58883e+07)(6.73315,3.56576e+07)(6.97978,3.54917e+07)(7.2338,3.54081e+07)(7.48377,3.53748e+07)(7.73168,3.53023e+07)(7.98013,3.51951e+07)(8.23273,3.51129e+07)(8.47922,3.4999e+07)(8.72585,3.48471e+07)(8.97312,3.46229e+07)(9.21884,3.43486e+07)(9.46557,3.41232e+07)(9.7168,3.38078e+07)(9.96301,3.34785e+07)(10.2101,3.32904e+07)(10.4565,3.323e+07)(10.7048,3.32134e+07)(10.9519,3.32104e+07)(11.2022,3.32081e+07)(11.4485,3.32095e+07)(11.6948,3.32095e+07)(11.942,3.32095e+07)(12.1886,3.32095e+07)(12.4359,3.32095e+07)(12.6829,3.32095e+07)(12.9298,3.32095e+07)(13.1769,3.32095e+07)(13.4242,3.32095e+07)(13.6714,3.32095e+07)(13.9177,3.32095e+07)(14.166,3.32095e+07)(14.4129,3.32095e+07)(14.6603,3.32095e+07)(14.9081,3.32095e+07)(15.155,3.32095e+07)(15.4018,3.32095e+07)(15.6488,3.32095e+07)(15.8998,3.32095e+07)(16.1436,3.32095e+07)(16.3878,3.32095e+07)(16.6378,3.32095e+07)(16.8847,3.32095e+07)(17.1331,3.32095e+07)(17.3828,3.32095e+07)(17.63,3.32095e+07)(17.8779,3.32095e+07)(18.1266,3.32095e+07)(18.3718,3.32095e+07)(18.6178,3.32095e+07)(18.8697,3.32095e+07)(19.117,3.32095e+07)(19.3668,3.32095e+07)(19.6134,3.32095e+07)(19.8592,3.32095e+07)(20.1057,3.32095e+07)(20.3519,3.32095e+07)(20.5995,3.32095e+07)(20.8456,3.32095e+07)(21.0914,3.32095e+07)(21.3381,3.32095e+07)(21.5844,3.32095e+07)(21.8308,3.32095e+07)(22.0767,3.32095e+07)(22.3226,3.32095e+07)(22.5704,3.32095e+07)(22.8155,3.32095e+07)(23.062,3.32095e+07)(23.3079,3.32095e+07)(23.5537,3.32095e+07)(23.8007,3.32095e+07)(24.0475,3.32095e+07)(24.2941,3.32095e+07)(24.5403,3.32095e+07)(24.7868,3.32095e+07)(25.0361,3.32095e+07)(25.2822,3.32095e+07)(25.5342,3.32095e+07)(25.7807,3.32095e+07)(26.0264,3.32095e+07)(26.2723,3.32095e+07)(26.5187,3.32095e+07)(26.7648,3.32095e+07)(27.0109,3.32095e+07)(27.259,3.32095e+07)(27.5056,3.32095e+07)(27.7519,3.32095e+07)(28.0043,3.32095e+07)(28.2509,3.32095e+07)(28.497,3.32095e+07)(28.7438,3.32095e+07)(28.991,3.32095e+07)(29.2374,3.32095e+07)(29.4843,3.32095e+07)(29.7315,3.32095e+07)(29.978,3.32095e+07)(30.2242,3.32095e+07)(30.4771,3.32095e+07)(30.7238,3.32095e+07)(30.9714,3.32095e+07)(31.2181,3.32095e+07)(31.4644,3.32095e+07)(31.7114,3.32095e+07)(31.9609,3.32095e+07)(32.2097,3.32095e+07)(32.4581,3.32095e+07)(32.7075,3.32095e+07)(32.9574,3.32095e+07)(33.2052,3.32095e+07)(33.4553,3.32095e+07)(33.7039,3.32095e+07)(33.9526,3.32095e+07)(34.2016,3.32095e+07)(34.4507,3.32095e+07)(34.7048,3.32095e+07)(34.9538,3.32095e+07)(35.2028,3.32095e+07)(35.4514,3.32095e+07)(35.7002,3.32095e+07)(35.9494,3.32095e+07)(36.1978,3.32095e+07)(36.4459,3.32095e+07)(36.6931,3.32095e+07)(36.9413,3.32095e+07)(37.1937,3.32095e+07)(37.4427,3.32095e+07)(37.6926,3.32095e+07)(37.9408,3.32095e+07)(38.1891,3.32095e+07)(38.4409,3.32095e+07)(38.6881,3.32095e+07)(38.9368,3.32095e+07)(39.186,3.32095e+07)(39.4343,3.32095e+07)(39.6834,3.32095e+07)(39.9326,3.32095e+07)(40.1873,3.32095e+07)(40.4355,3.32095e+07)(40.6835,3.32095e+07)(40.9336,3.32095e+07)(41.182,3.32095e+07)(41.4309,3.32095e+07)(41.6797,3.32095e+07)(41.9253,3.32095e+07)(42.1747,3.32095e+07)(42.4217,3.32095e+07)(42.6707,3.32095e+07)(42.9198,3.32095e+07)(43.1689,3.32095e+07)(43.4184,3.32095e+07)(43.6669,3.32095e+07)(43.9164,3.32095e+07)(44.1666,3.32095e+07)(44.4155,3.32095e+07)(44.6644,3.32095e+07)(44.9134,3.32095e+07)(45.163,3.32095e+07)(45.4122,3.32095e+07)(45.6614,3.32095e+07)(45.9099,3.32095e+07)(46.1587,3.32095e+07)(46.4081,3.32095e+07)(46.6577,3.32095e+07)(46.9068,3.32095e+07)(47.1553,3.32095e+07)(47.4038,3.32095e+07)(47.6531,3.32095e+07)(47.9027,3.32095e+07)(48.1518,3.32095e+07)(48.4004,3.32095e+07)(48.6482,3.32095e+07)(48.8972,3.32095e+07)(49.1459,3.32095e+07)(49.3981,3.32095e+07)(49.6476,3.32095e+07)(49.8965,3.32095e+07)(50.1466,3.32095e+07)(50.3947,3.32095e+07)(50.6436,3.32095e+07)(50.8927,3.32095e+07)(51.1412,3.32095e+07)(51.3905,3.32095e+07)(51.6395,3.32095e+07)(51.8912,3.32095e+07)(52.1487,3.32095e+07)(52.4011,3.32095e+07)(52.6495,3.32095e+07)(52.8998,3.32095e+07)(53.1491,3.32095e+07)(53.3978,3.32095e+07)(53.646,3.32095e+07)(53.8955,3.32095e+07)(54.1438,3.32095e+07)(54.3938,3.32095e+07)(54.6425,3.32095e+07)(54.8911,3.32095e+07)(55.1406,3.32095e+07)(55.3903,3.32095e+07)(55.6387,3.32095e+07)(55.887,3.32095e+07)(56.1372,3.32095e+07)(56.3861,3.32095e+07)(56.6361,3.32095e+07)(56.8847,3.32095e+07)(57.1335,3.32095e+07)(57.3833,3.32095e+07)(57.6324,3.32095e+07)(57.8811,3.32095e+07)(58.1301,3.32095e+07)(58.3791,3.32095e+07)(58.6309,3.32095e+07)(58.8793,3.32095e+07)(59.1284,3.32095e+07)(59.3778,3.32095e+07)(59.626,3.32095e+07)(59.8748,3.32095e+07)
	 	};
	 	\addlegendentry{QP\textsubscript{clique}}
	 	\addplot[
	 	color=blue,
	 	mark=cirlce,
	 	]
	 	coordinates {
	 		(0,1.07137e+08)(21.7831,2.27839e+07)(43.529,1.97244e+07)(65.6666,1.90742e+07)
	 		
	 	};
	 	\addlegendentry{LP\textsubscript{clique}}
	 	
	 	\end{axis}
	 	\end{tikzpicture}\\
	 		\begin{tikzpicture}[scale=0.85]
	 	\centering %image 4_3_s was used
	 	\begin{axis}[
	 	title = {Dense CRF},
	 	xlabel = {Time (s)},
	 	ylabel = {Energy},
	 	xmin = {0},
	 	xmax = {60},
	 	legend pos = north east,]
	 	]
	 	\addplot[
	 	color=green,
	 	mark=cirlce,
	 	]
	 	coordinates {
	 		(0,9.74179e+07)(0.050173,8.96315e+07)(0.09738,8.39752e+07)(0.184372,7.86267e+07)(0.265161,7.33083e+07)(10,7.33083e+07)(20,7.33083e+07)(30,7.33083e+07)(40,7.33083e+07)(50,7.33083e+07)(60,7.33083e+07)
	 		
	 	};	
 		\addlegendentry{{MF5}}
	 	\addplot[
	 	color=red,
	 	mark=cirlce,
	 	]
	 	coordinates {
	 		(0,1.04608e+08)(0.058598,9.64109e+07)(0.113172,8.94973e+07)(0.167933,8.37345e+07)(0.222721,7.83544e+07)(0.277284,7.32363e+07)(0.331863,6.89079e+07)(0.3864,6.50192e+07)(0.440994,6.22147e+07)(0.495716,6.02109e+07)(0.550018,5.8751e+07)(0.604505,5.73647e+07)(0.659065,5.6133e+07)(0.713792,5.49229e+07)(0.768267,5.41196e+07)(0.822771,5.34116e+07)(0.877273,5.2807e+07)(0.931915,5.22459e+07)(0.986521,5.16327e+07)(1.04092,5.10128e+07)(1.09541,5.04245e+07)(1.14985,4.97518e+07)(1.20454,4.90502e+07)(1.25945,4.83303e+07)(1.31399,4.77316e+07)(1.36848,4.72023e+07)(1.42305,4.67233e+07)(1.47841,4.63575e+07)(1.53391,4.60211e+07)(1.58958,4.576e+07)(1.64391,4.56499e+07)(1.69816,4.55706e+07)(1.75281,4.54215e+07)(1.80731,4.52868e+07)(1.86186,4.51352e+07)(1.91632,4.49774e+07)(1.97089,4.4808e+07)(2.02541,4.46724e+07)(2.07979,4.45851e+07)(2.13406,4.44852e+07)(2.18875,4.43321e+07)(2.24319,4.41587e+07)(2.29791,4.39065e+07)(2.35266,4.36789e+07)(2.40735,4.34592e+07)(2.46176,4.32607e+07)(2.51631,4.31351e+07)(2.57065,4.30267e+07)(2.62521,4.29843e+07)(2.67996,4.29693e+07)(2.73884,4.29693e+07)(10,4.29693e+07)(20,4.29693e+07)(30,4.29693e+07)(40,4.29693e+07)(50,4.29693e+07)(60,4.29693e+07)
	 		
	 	};
	 	\addlegendentry{QP}
	 	\addplot[
	 	color=blue,
	 	mark=cirlce,
	 	]
	 	coordinates {
	 		(0,1.04608e+08)(5.7476,2.91798e+07)(11.5448,1.93042e+07)(17.3227,1.69672e+07)(23.0966,1.57283e+07)(28.8789,1.4974e+07)(34.6514,1.44655e+07)(40.4509,1.41802e+07)(46.217,1.38941e+07)(51.975,1.35936e+07)(57.7285,1.33068e+07)(63.4511,1.29287e+07)(69.1339,1.24849e+07)(74.84,1.19422e+07)(80.5338,1.15031e+07)(86.2791,1.12581e+07)(92.0388,1.09056e+07)(97.7789,1.04758e+07)(103.496,9.89856e+06)(109.272,9.6684e+06)(115.03,9.39995e+06)
	 	};
	 	\addlegendentry{LP}
	 	
	 	\end{axis}
	 	\end{tikzpicture}
	 	\begin{tikzpicture}[scale=0.85]
	 	\centering %image 8_19_s was used
	 	\begin{axis}[
	 	title = {Dense CRF with high-order terms},
	 	xlabel = {Time (s)},
	 	ylabel = {Energy},
	 	xmin = {0},
	 	xmax = {50},
	 	legend pos = north east,]
	 	]
	 	\addplot[
	 	color=green,
	 	mark=cirlce,
	 	]
	 	coordinates {
	 		(0,4.32354e+08)(0.070828,3.6835e+08)(0.133018,3.2554e+08)(0.23535,2.93865e+08)(0.330325,2.67015e+08)(10,2.67015e+08)(20,2.67015e+08)(30,2.67015e+08)(40,2.67015e+08)(50,2.67015e+08)(60,2.67015e+08)
	 		
	 	};
	 	\addlegendentry{MF5\textsubscript{clique}}
	 		
	 	\addplot[
	 	color=blue,
	 	mark=cirlce,
	 	]
	 	coordinates {
	 		(0,4.83714e+08)(5.88663,1.10999e+08)(11.7032,9.33048e+07)(17.4946,8.08034e+07)(23.2975,7.48701e+07)(29.1005,7.14922e+07)(34.8887,6.90398e+07)(40.6874,6.76615e+07)(46.4888,6.59784e+07)(52.2838,6.45633e+07)(58.0758,6.31786e+07)(63.8593,6.18749e+07)(69.6458,6.04648e+07)
	 		
	 	};
	 	\addlegendentry{LP\textsubscript{clique}}
	 	
	 	\addplot[
	 	color=red,
	 	mark=cirlce,
	 	]
	 	coordinates {
	 		(0,4.83714e+08)(0,4.26234e+08)(0.093691,3.77567e+08)(0.184382,3.35275e+08)(0.275122,3.00331e+08)(0.366082,2.71814e+08)(0.456407,2.51472e+08)(0.547176,2.37259e+08)(0.637706,2.26638e+08)(0.728447,2.17564e+08)(0.819609,2.10325e+08)(0.91025,2.03829e+08)(1.00092,1.98096e+08)(1.09187,1.92756e+08)(1.18273,1.88477e+08)(1.27332,1.84217e+08)(1.36425,1.80236e+08)(1.45529,1.76763e+08)(1.54612,1.73038e+08)(1.63713,1.6903e+08)(1.72792,1.65554e+08)(1.81866,1.62259e+08)(1.90955,1.59062e+08)(2.0004,1.56181e+08)(2.09101,1.53626e+08)(2.18185,1.5126e+08)(2.27274,1.49025e+08)(2.36401,1.47151e+08)(2.4549,1.45659e+08)(2.54549,1.44333e+08)(2.63609,1.43494e+08)(2.72699,1.43159e+08)(2.81755,1.42845e+08)(2.90848,1.42746e+08)(2.99909,1.42641e+08)(3.08979,1.42546e+08)(3.18064,1.42449e+08)(3.27146,1.4237e+08)(3.36195,1.42274e+08)(3.4527,1.42237e+08)(3.5435,1.42201e+08)(3.63419,1.4219e+08)(3.72509,1.42157e+08)(3.81591,1.42139e+08)(3.90651,1.42127e+08)(3.99708,1.42127e+08)(4.08773,1.42127e+08)(10,1.42127e+08)(20,1.42127e+08)(30,1.42127e+08)(40,1.42127e+08)(50,1.42127e+08)(60,1.42127e+08)
	 	};
	 	\addlegendentry{QP\textsubscript{clique}}
	 	
	 	\end{axis}
	 	\end{tikzpicture}
	 \end{tabular}
	 	
	 	\caption{\textit{Assignment energies for the sheep image from Pascal (top row) and small plane image from MSRC (bottom row) as a function of time when the parameters are tuned to \textbf{QP\textsubscript{clique}} and \textbf{QP}. The left graphs show the assignment energy calculated using only the unary and pairwise potentials, the right image shows the assignment energy of (\ref{eq:sp_ip}), which consists of the unary, pairwise and higher-order potentials. It is worth noting the first iteration of \textbf{LP\textsubscript{clique}} and \textbf{LP}, achieves a lower energy then the final energy of \textbf{QP\textsubscript{clique}} and \textbf{QP} respectively, further highlighting the sophistication of the LP minimisation.}}\label{fig:qp_energies}
	 \end{figure}
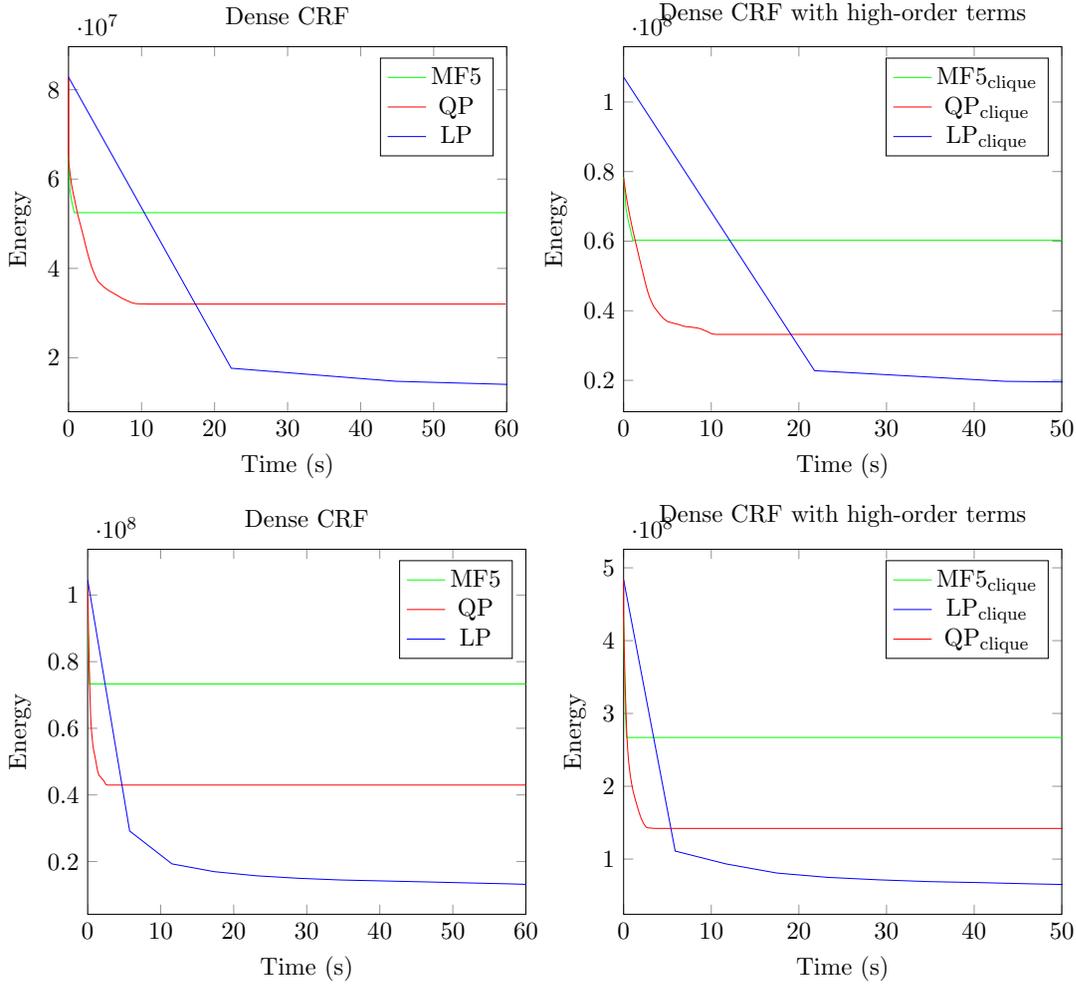

\begin{figure*}
	\def \SUBWIDTH{0.12\linewidth}
	\begin{center}		
		\begin{subfigure}{\SUBWIDTH}
			\includegraphics[width=0.99\linewidth]{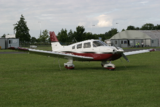}
			%					\caption{Image}
		\end{subfigure}%
		\begin{subfigure}{\SUBWIDTH}
			\includegraphics[width=0.99\linewidth]{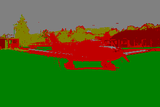}
			%				\caption{MF5}
		\end{subfigure}%
		\begin{subfigure}{\SUBWIDTH}
			\includegraphics[width=0.99\linewidth]{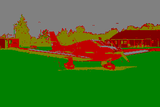}
			%			\caption{QP}
		\end{subfigure}%
		\begin{subfigure}{\SUBWIDTH}
			\includegraphics[width=0.99\linewidth]{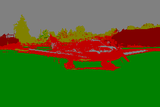}
			%					\caption{LP}
		\end{subfigure}%
		\begin{subfigure}{\SUBWIDTH}
			\includegraphics[width=0.99\linewidth]{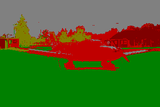}
			%					\caption{MF5}
		\end{subfigure}%
		\begin{subfigure}{\SUBWIDTH}
			\includegraphics[width=0.99\linewidth]{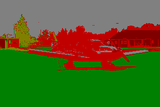}
			%					\caption{QP}
		\end{subfigure}%
		\begin{subfigure}{\SUBWIDTH}
			\includegraphics[width=0.99\linewidth]{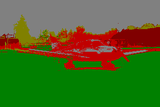}
			%					\caption{LP}
		\end{subfigure}%
		\begin{subfigure}{\SUBWIDTH}
			\includegraphics[width=0.99\linewidth]{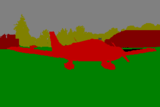}
			%					\caption{GT}
		\end{subfigure}
		
			\begin{subfigure}{\SUBWIDTH}
				\includegraphics[width=0.99\linewidth]{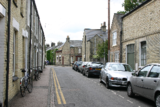}
				%					\caption{Image}
			\end{subfigure}%
			\begin{subfigure}{\SUBWIDTH}
				\includegraphics[width=0.99\linewidth]{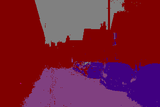}
				%				\caption{MF5}
			\end{subfigure}%
			\begin{subfigure}{\SUBWIDTH}
				\includegraphics[width=0.99\linewidth]{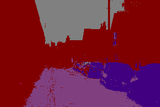}
				%			\caption{QP}
			\end{subfigure}%
			\begin{subfigure}{\SUBWIDTH}
				\includegraphics[width=0.99\linewidth]{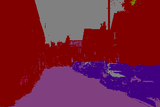}
				%					\caption{LP}
			\end{subfigure}%
			\begin{subfigure}{\SUBWIDTH}
				\includegraphics[width=0.99\linewidth]{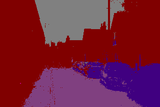}
				%					\caption{MF5}
			\end{subfigure}%
			\begin{subfigure}{\SUBWIDTH}
				\includegraphics[width=0.99\linewidth]{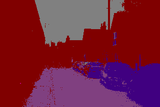}
				%					\caption{QP}
			\end{subfigure}%
			\begin{subfigure}{\SUBWIDTH}
				\includegraphics[width=0.99\linewidth]{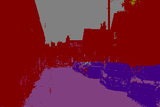}
				%					\caption{LP}
			\end{subfigure}%
			\begin{subfigure}{\SUBWIDTH}
				\includegraphics[width=0.99\linewidth]{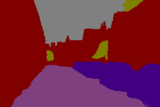}
				%					\caption{GT}
			\end{subfigure}
	
				\begin{subfigure}{\SUBWIDTH}
					\includegraphics[width=0.99\linewidth]{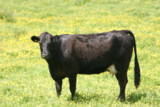}
					%					\caption{Image}
				\end{subfigure}%
				\begin{subfigure}{\SUBWIDTH}
					\includegraphics[width=0.99\linewidth]{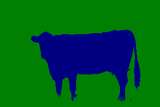}
					%				\caption{MF5}
				\end{subfigure}%
				\begin{subfigure}{\SUBWIDTH}
					\includegraphics[width=0.99\linewidth]{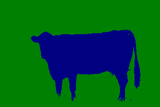}
					%			\caption{QP}
				\end{subfigure}%
				\begin{subfigure}{\SUBWIDTH}
					\includegraphics[width=0.99\linewidth]{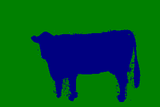}
					%					\caption{LP}
				\end{subfigure}%
				\begin{subfigure}{\SUBWIDTH}
					\includegraphics[width=0.99\linewidth]{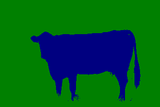}
					%					\caption{MF5}
				\end{subfigure}%
				\begin{subfigure}{\SUBWIDTH}
					\includegraphics[width=0.99\linewidth]{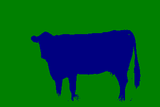}
					%					\caption{QP}
				\end{subfigure}%
				\begin{subfigure}{\SUBWIDTH}
					\includegraphics[width=0.99\linewidth]{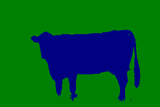}
					%					\caption{LP}
				\end{subfigure}%
				\begin{subfigure}{\SUBWIDTH}
					\includegraphics[width=0.99\linewidth]{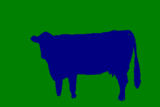}
					%					\caption{GT}
				\end{subfigure}

		%% pascal
		
		\begin{subfigure}{\SUBWIDTH}
			\includegraphics[width=0.99\linewidth]{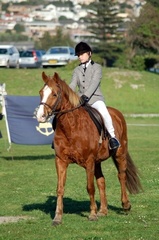}
			%					\caption{Image}
		\end{subfigure}%
		\begin{subfigure}{\SUBWIDTH}
			\includegraphics[width=0.99\linewidth]{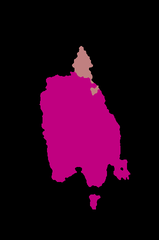}
			%				\caption{MF5}
		\end{subfigure}%
		\begin{subfigure}{\SUBWIDTH}
			\includegraphics[width=0.99\linewidth]{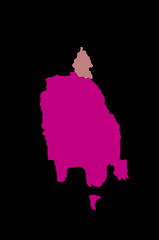}
			%			\caption{QP}
		\end{subfigure}%
		\begin{subfigure}{\SUBWIDTH}
			\includegraphics[width=0.99\linewidth]{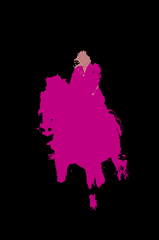}
			%					\caption{LP}
		\end{subfigure}%
		\begin{subfigure}{\SUBWIDTH}
			\includegraphics[width=0.99\linewidth]{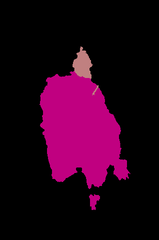}
			%					\caption{MF5}
		\end{subfigure}%
		\begin{subfigure}{\SUBWIDTH}
			\includegraphics[width=0.99\linewidth]{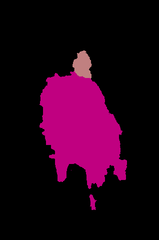}
			%					\caption{QP}
		\end{subfigure}%
		\begin{subfigure}{\SUBWIDTH}
			\includegraphics[width=0.99\linewidth]{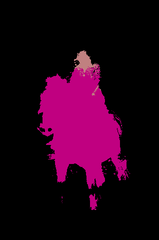}
			%					\caption{LP}
		\end{subfigure}%
		\begin{subfigure}{\SUBWIDTH}
			\includegraphics[width=0.99\linewidth]{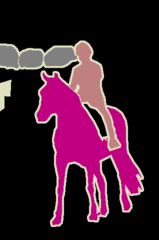}
			%					\caption{GT}
		\end{subfigure}
		
		\begin{subfigure}{\SUBWIDTH}
			\includegraphics[width=0.99\linewidth]{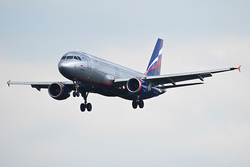}
			%					\caption{Image}
		\end{subfigure}%
		\begin{subfigure}{\SUBWIDTH}
			\includegraphics[width=0.99\linewidth]{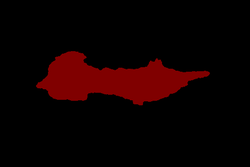}
			%				\caption{MF5}
		\end{subfigure}%
		\begin{subfigure}{\SUBWIDTH}
			\includegraphics[width=0.99\linewidth]{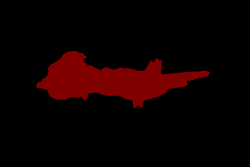}
			%			\caption{QP}
		\end{subfigure}%
		\begin{subfigure}{\SUBWIDTH}
			\includegraphics[width=0.99\linewidth]{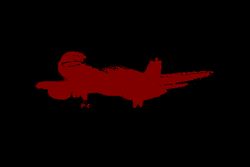}
			%					\caption{LP}
		\end{subfigure}%
		\begin{subfigure}{\SUBWIDTH}
			\includegraphics[width=0.99\linewidth]{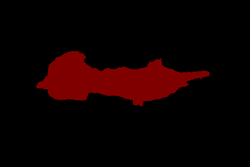}
			%					\caption{MF5}
		\end{subfigure}%
		\begin{subfigure}{\SUBWIDTH}
			\includegraphics[width=0.99\linewidth]{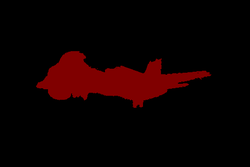}
			%					\caption{QP}
		\end{subfigure}%
		\begin{subfigure}{\SUBWIDTH}
			\includegraphics[width=0.99\linewidth]{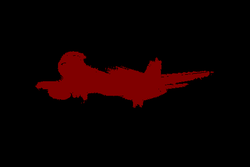}
			%					\caption{LP}
		\end{subfigure}%
		\begin{subfigure}{\SUBWIDTH}
			\includegraphics[width=0.99\linewidth]{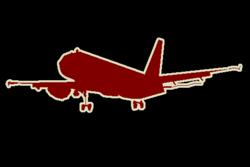}
			%					\caption{GT}
		\end{subfigure}

		\begin{subfigure}{\SUBWIDTH}
			\includegraphics[width=0.99\linewidth]{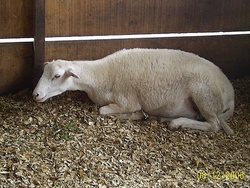}
			\caption*{Image}
		\end{subfigure}%
		\begin{subfigure}{\SUBWIDTH}
			\includegraphics[width=0.99\linewidth]{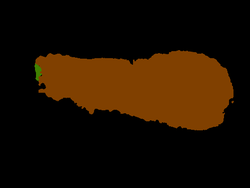}
			\caption*{\textbf{MF5}}
		\end{subfigure}%
		\begin{subfigure}{\SUBWIDTH}
			\includegraphics[width=0.99\linewidth]{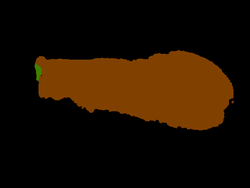}
			\caption*{\textbf{QP}}
		\end{subfigure}%
		\begin{subfigure}{\SUBWIDTH}
			\includegraphics[width=0.99\linewidth]{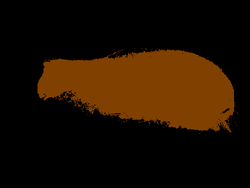}
			\caption*{\textbf{LP}}
		\end{subfigure}%
		\begin{subfigure}{\SUBWIDTH}
			\includegraphics[width=0.99\linewidth]{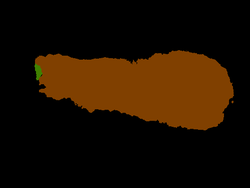}
			\caption*{\textbf{MF5}\textsubscript{clique}}
		\end{subfigure}%
		\begin{subfigure}{\SUBWIDTH}
			\includegraphics[width=0.99\linewidth]{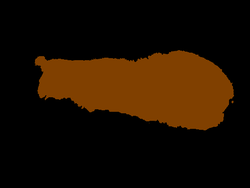}
			\caption*{\textbf{QP\textsubscript{clique}}}
		\end{subfigure}%
		\begin{subfigure}{\SUBWIDTH}
			\includegraphics[width=0.99\linewidth]{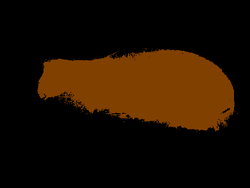}
			\caption*{\textbf{LP}\textsubscript{clique}}
		\end{subfigure}%
		\begin{subfigure}{\SUBWIDTH}
			\includegraphics[width=0.99\linewidth]{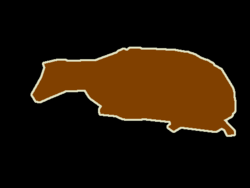}
			\caption*{GT}
		\end{subfigure}
		
		\vspace{-0.2cm}
		\caption{\em Qualitative results for MSRC (top three) and Pascal (bottom three) with the parameters
			tuned for \textbf{QP\textsubscript{clique}} and \textbf{QP}. For Pascal, \textbf{LP} achieves to most accurate segmentations, however, this is not the case for MSRC.}
		\label{fig:qp_images}
	\end{center}
	\vspace{-0.9cm}
\end{figure*}

	 In order to compare energy values, for \textbf{QP}, \textbf{LP} and \textbf{MF5} we used the parameters tuned to \textbf{QP}. For \textbf{QP\textsubscript{clique}}, \textbf{LP\textsubscript{clique}} and  \textbf{MF5\textsubscript{clique}} the parameters were tuned to \textbf{QP\textsubscript{clique}}, Table \ref{tab:qp_energies} gives the results for all algorithms and Figure \ref{fig:qp_energies} shows a decrease in energy at runtime. Whilst \textbf{LP} and \textbf{LP\textsubscript{clique}} could be initialised with a faster algorithm such as \textbf{QP}, we chose to presents the results in an ``as is" fashion, the interested reader is encouraged to visit \cite{EfficientLinearProgrammingforDenseCRFs} for an example of when \textbf{LP} is initialised with a faster algorithm. Qualitative methods can be seen in Figure \ref{fig:qp_images}.
	 
	 %%% accuracy table %%%
	 \begin{table}[h]\centering
	 	\begin{tabular}{@{}ccccc@{}}
	 		\toprule
	 		Algorithm                         & \textbf{Avg.E ($\times10^7$)} & \textbf{Time(s)} & \textbf{Acc(\%)} & \textbf{IoU(\%)} \\\hline
	 		\multicolumn{5}{c}{Pascal} \\\hline
	 		\textbf{MF5}                      & 2.92  & \textbf{0.7} & 79.42 & 22.21 \\
	 		\textbf{QP}                       & 0.97  & 9.8 & {79.51} & 22.19\\
	 		\textbf{LP}                       & \textbf{0.62}  & 236.5 & \textbf{79.84} & 21.80\\\cline{1-2}
	 		\textbf{MF5\textsubscript{clique}} & 5.20  & 1.2 & 79.44 & {22.22}\\	 	
	 		\textbf{QP\textsubscript{clique}} & 3.78  & 12.4 & 79.54 & 22.21\\	 		
	 		\textbf{LP\textsubscript{clique}} & \textbf{2.19}  & 254.2 & 79.80 & \textbf{22.22}\\
	 		
	 		\toprule
	 		\multicolumn{5}{c}{MSRC} \\\hline
 	 		  	 		\textbf{MF5}                      & 58.9  & \textbf{0.27} & 83.79 & 57.16 \\
 	 		  	 		\textbf{QP}                       & 29.2  & 1.06 & \textbf{83.93} & 57.80\\
 	 		  	 		\textbf{LP}                       & \textbf{13.8}  & 54.0 & 82.93 & 57.30\\\cline{1-2}
 	 		  	 		\textbf{MF5\textsubscript{clique}} & 73.6  & 0.475 & 83.404 & {57.81}\\	
 	 		  	 		\textbf{QP\textsubscript{clique}} & 46.1  & 1.75 & 83.56 & \textbf{57.81}\\	 		
 	 		  	 		\textbf{LP\textsubscript{clique}} & \textbf{44.1}  & 49.3 & 81.49 & 55.81\\
	 		\bottomrule
	 	\end{tabular}\caption{\textit{Table displaying the average energy, timings, accuracy and IoU, when the parameters are tuned to \textbf{QP\textsubscript{clique}} and \textbf{QP}. It is shown that the lowest energies are achieved by \textbf{LP\textsubscript{clique}} and \textbf{LP}.Interestingly the inclusion of higher-order terms reduces the pixel accuracy but provides a slight increase in IoU score.}\label{tab:qp_energies}}
	 \end{table}

%%% segmentation images %%%
	  
	 The results given in Table  \ref{tab:qp_energies} clearly show that the LP relaxations achieve lower energies when compared to their QP counterparts, this is not surprising as the LP relaxation used in this paper is known to give a tighter relaxation then QP \cite{Kumar08ananalysis}. As is consistent with our previous work \cite{EfficientLinearProgrammingforDenseCRFs},~\textbf{LP} also obtains lower energies then \textbf{MF5}. For consistency we also performed a set of experiments with the parameters tuned to \textbf{MF5\textsubscript{clique}} and \textbf{MF5}, given in Appendix \ref{app:res}. In this setting, the same pattern is observed where \textbf{LP\textsubscript{clique}} achieves lower energies then its \textbf{QP} and \textbf{MF5} counterparts. 
	 In summary, \textbf{QP\textsubscript{clique}} achieves fast initial energy minimisation but converges to a local minimum and fails to reach the low energies achieved by \textbf{LP\textsubscript{clique}}. However, whilst using continuous relaxations clearly achieves lower energies, it is not apparent as to whether continuous relaxations improves segmentation accuracy. Similarly to our previous works \cite{EfficientContinuousRelaxationsforDenseCRF, EfficientLinearProgrammingforDenseCRFs}, the segmentation performance is heavily dependant on the choice of parameters. As such, further work would include investigating the learning of such parameters, possibly in a deep-learning setting as in \cite{zheng2015conditional}.

\section{Discussion}
The primary contributions of this paper are a quadratic programming and a linear programming relaxation for minimising a dense CRF with sparse higher-order potentials. Due to the use of Gaussian Pairwise potentials and enforcing labelling consistency in the higher-order terms, each iteration of both algorithms exhibit a time complexity linear in the number of labels and pixels. It is the tightness of the relaxations coupled with the sophistication of the optimisation techniques that allows both approaches to achieve lower energies than state-of-the-art methods. Further work would include incorporating the methods into an end-to-end learning framework \cite{zheng2015conditional}, which would focus on achieving accurate segmentation results as well as low energies.

\clearpage
 	  		  
		  \appendix\nopunct
		  \section{}		  
		  \subsection{Optimal Step Size for the Frank-Wolfe algorithm}\label{app:step_size}
		  For an efficient Frank-Wolfe algorithm, an optimal step size is essential and forms one of the three key steps defined in Algorithm \ref{algo:fw}. This section details how the optimal step size is calculated, the optimal step size to the Frank-Wolfe algorithm is achieved by solving:
		  \begin{align}\label{eq:app_step_size_calc}
		  \delta^* = \argmin_{\delta \in [0,1]}{f(\mathbf{y} + \delta(\mathbf{s_{y}} - \mathbf{y}),\mathbf{z} + \delta(\mathbf{s_{z}} - \mathbf{z}))},
		  \end{align}
		  where:
		  \begin{align}
			\min_\mathbf{y,z}{f(\mathbf{y},\mathbf{z})} &= \min_\mathbf{y,z}\big(\boldsymbol{\phi}^T\mathbf{y} + \mathbf{y}^T\boldsymbol{\varPsi}\mathbf{y} + 
\mathbf{c}^T\mathbf{z} + \mathbf{(1_z - z)}^T\mathbf{C}\boldsymbol{H}\mathbf{(1_y - y)}\big).
		  \end{align}
		  Expanding out equation (\ref{eq:app_step_size_calc}) and collecting terms of $\delta$ yields:
		  \begin{equation}		 
		  \begin{split}
		  \delta^* = \argmin_{\delta \in [0,1]} \Bigg(
		  \delta^2\Big(&\mathbf{(s_{y} - y)^\textit{T}\boldsymbol{\varPsi}{(s_{y} - y)} + (s_{z} - z)^\textit{T}\mathbf{C}\boldsymbol{H}(s_{y} - y)}\Big) \\ + 
		  \delta\Big(&\mathbf{\boldsymbol{\phi}^\textit{T}(s_{y} - y) + 2(s_{y} - y)^\textit{T}\boldsymbol{\varPsi}{y} + \mathbf{c}^\textit{T}(s_{z} - z)} \\  - &\mathbf{(1_z - z)^\textit{T}\mathbf{C}\boldsymbol{H}(s_{y} - y) - (s_{z} - z)^\textit{T}\mathbf{C}\boldsymbol{H}(1_y - y)\Big)} \\ +
		  \Big(&\mathbf{\boldsymbol{\phi}^\textit{T}y + y^\textit{T}\boldsymbol{\varPsi}y + \mathbf{c}^\textit{T}\mathbf{z}  + (1_z - z)^\textit{T}\mathbf{C}\boldsymbol{H}(1_y - y)}\Big)\Bigg).
		  \end{split}	
		  \end{equation}
		  This equation is quadratic in $\delta$ and hence the minimum value has a closed form solution given as:
		  \begin{equation}\label{eq:optimal_step}
		  \begin{split}
		  \delta^* =P_{[0,1]}\Bigg[ &- \frac{1}{2}\frac{\mathbf{\boldsymbol{\phi}^\textit{T}(s_{y} - y) + 2(s_{y} - y)^\textit{T}\boldsymbol{\varPsi}{y} + \mathbf{c}^\textit{T}(s_{z} - z)}}{\mathbf{(s_{y} - y)^\textit{T}\boldsymbol{\varPsi}{(s_{y} - y)} + (s_{z} - z)^\textit{T}\mathbf{C}\boldsymbol{H}(s_{y} - y)}} \\ & + \frac{1}{2}\frac{\mathbf{(1_z - z)^\textit{T}\mathbf{C}\boldsymbol{H}(s_{y} - y) - (s_{z} - z)^\textit{T}\mathbf{C}\boldsymbol{H}(1_y - y)}}{\mathbf{(s_{y} - y)^\textit{T}\boldsymbol{\varPsi}{(s_{y} - y)} + (s_{z} - z)^\textit{T}\mathbf{C}\boldsymbol{H}(s_{y} - y)}}\Bigg],
		  \end{split}
		  \end{equation}
		  and for the Frank-Wolfe algorithm, is the optimal step size. $P_{[0,1]}$ indicates that if the value of $\delta^*$ falls outside of the range $[0,1]$, then the optimal step size is truncated to lie within this range.
		  
		  \subsection{Formulation of the Lagrange Dual for the LP}\label{app:lp_dual}
		  Starting with the primal problem, which is given as:
		  \begin{align}
		  \min_\mathbf{y,v,w} \sum_{a=1}^N \sum_{i \in\cal{L}}&\phi_{a:i}{y_{a:i}} + \sum_{a=1}^N\sum_{\substack{b=1 \\ b \neq a}}^N\sum_{i \in \mathcal{L}}\frac{\text{K}_{ab}}{2}v_{ab:i} + \sum_{p=1}^RC_pw_p + \frac{1}{2\lambda}||\mathbf{y} - \mathbf{y}^k||^2,\\\nonumber
		  \text{s.t  } v_{ab:i} &\geq y_{a:i} - y_{b:i} \indent\forall a,b \in \{1,...,N\} \indent a \neq b \indent\forall i \in \mathcal{L},  \\\nonumber
		  v_{ab:i} & \geq y_{b:i} - y_{a:i}\indent\forall a,b \in \{1,...,N\} \indent a \neq b \indent\forall i \in \mathcal{L},  \\\nonumber
		  w_p & \geq y_{c:pi} - y_{d:pi}\quad\forall c,d \in \mathcal{R}_p \indent c \neq d \indent\forall i \in \mathcal{L}\indent\forall p \in \{1,...,R\},  \\\nonumber
		  w_p & \geq y_{d:pi} - y_{c:pi}\quad\forall c,d \in \mathcal{R}_p \indent c \neq d \indent\forall i \in \mathcal{L}\indent\forall p \in \{1,...,R\}, \\\nonumber
		  y_{a:i}  & \geq 0\text{\space}\indent\indent\indent\forall a \in \{1,...,N\} \indent\forall i \in \mathcal{L},\\\nonumber
		  \sum_{i \in \mathcal{L}}y_{a:i}  &= 1\text{\space}\indent\indent\indent\forall a \in \{1,...,N\}.\nonumber
		  \end{align}
		 The associated Lagrangian can thus be written as:
		  \begin{align}\label{eq:lp_lagrangian}
		  \max_{\boldsymbol{\alpha},\boldsymbol{\mu},\boldsymbol{\beta},\boldsymbol{\gamma}}&\min_\mathbf{y,w,v} L({\boldsymbol{\alpha},\boldsymbol{\mu},\boldsymbol{\beta},\boldsymbol{\gamma}},\mathbf{y,w,v})  = \\\nonumber&\sum_{a=1}^N\sum_{i \in \mathcal{L}}\phi_{a:i}{y_{a:i}} + \sum_{a=1}^N\sum_{\substack{b=1 \\ b \neq a}}^N\sum_{i \in \mathcal{L}}\frac{\text{K}_{ab}}{2}v_{ab:i} + \sum_{p=1}^RC_pw_p + \frac{1}{2\lambda}||\mathbf{y} - \mathbf{y}^k|| \\\nonumber& 
		  - \sum_{a=1}^N\sum_{\substack{b=1 \\ b \neq a}}^N\sum_{i \in \mathcal{L}}\alpha^1_{ab:i}(y_{b:i} - y_{a:i} + v_{ab:i}) - \sum_{a=1}^N\sum_{\substack{b=1 \\ b \neq a}}^N\sum_{i \in \mathcal{L}}\alpha^2_{ab:i}(y_{a:i} - y_{b:i} + v_{ab:i})\nonumber\\&
		  - \sum_{p=1}^R\sum_{\substack{c,d \in \mathcal{R}_p \\\nonumber c\neq d }}\sum_{i \in \mathcal{L}}\mu^1_{cd:pi}(y_{d:i} - y_{c:i} + w_p) - \sum_{p=1}^R\sum_{\substack{c,d \in \mathcal{R}_p \\ c\neq d }}\sum_{i \in \mathcal{L}}\mu^2_{cd:pi}(y_{c:i} - y_{d:i} + w_p)\\\nonumber&
		  + \sum_a\beta_a\Big(1 - \sum_{i \in \mathcal{L}}y_{a:i}\Big) - \sum_a\sum_{i \in \mathcal{L}}\gamma_{a:i}y_{a:i}, \\\nonumber
		  \text{s.t}\indent \alpha^1_{ab:i},\alpha^2_{ab:i} & \geq 0 \indent \forall a,b \in \{1,...,N\} \indent a \neq b \indent\forall i \in \mathcal{L},\\\nonumber
		  \mu^1_{cd:pi},\mu^2_{cd:i} & \geq 0 \indent\forall c,d \in \mathcal{R}_p \indent c \neq d \indent\forall i \in \mathcal{L}\indent\forall p \in \{1,...,R\},\\\nonumber
		  \gamma_{a:i} & \geq 0 \indent \forall a \in \{1,...,N\}  \indent \forall i \in \mathcal{L}.
		  \end{align}
		  Here $\alpha^1_{ab:i}$, $\alpha^2_{ab:i}$, $\mu^1_{cd:pi}$, $\mu^2_{cd:pi}$, $\beta_a$ and $\gamma_{a:i}$ are the Lagrange variables. To obtain the dual problem, the Lagrangian needs to be minimised over the primal variables $\mathbf{y,w,v}$. When the derivatives of the Lagrangian with respect to $\mathbf{w}$ and $\mathbf{v}$ are non-zero, the problem is unbounded and hence the minimisation yields a value of $-\infty$. To this extent, the derivatives of the Lagrangian with respect to $\mathbf{w}$ and $\mathbf{v}$ must be zero for a bounded solution. These conditions are instrumental in obtaining constraints on the Lagrange multipliers $\alpha_{ab:i}^1,\alpha_{ab:i}^2$ and $\mu_{cd:pi}^1,\mu_{cd:pi}^2$. By rearranging $\nabla_\mathbf{v}L({\boldsymbol{\alpha},\boldsymbol{\mu},\boldsymbol{\beta},\boldsymbol{\gamma}},\mathbf{y,w,v}) = 0$ and $\nabla_L\mathbf{w}({\boldsymbol{\alpha},\boldsymbol{\mu},\boldsymbol{\beta},\boldsymbol{\gamma}},\mathbf{y,w,v}) = 0$ the constraints for the Lagrange multipliers $\alpha_{ab:i}^1,\alpha_{ab:i}^2$ and $\mu_{cd:pi}^1,\mu_{cd:pi}^2$ are obtained and given respectively as:
		  \begin{align}
		  \label{eq:constraint_pair}\alpha_{ab:i}^1 + \alpha_{ab:i}^2 & = \frac{K_{ab}}{2} \indent \forall a,b \in \{1,...,N\} \indent a \neq b \indent\forall i \in \mathcal{L}, \\
		  \label{eq:constraint_clique_lp}\sum_{i \in \mathcal{L}}\sum_{c,d}\mu_{cd:pi}^1 + \mu_{cd:pi}^2 & = C_{p  } \qquad\text{\space} \forall c,d \in \mathcal{R}_p \indent c \neq d \indent\forall i \in \mathcal{L}.
		  \end{align}
		  By differentiating the Lagrangian with respect to $\mathbf{y}$ and setting the derivative to zero an equation for the primal variables can be obtained. Before solving $\nabla_\mathbf{y}L({\boldsymbol{\alpha},\boldsymbol{\mu},\boldsymbol{\beta},\boldsymbol{\gamma}},\mathbf{y,w,v}) = 0$, it is beneficial to reorder the terms in the Lagrangian, using equations (\ref{eq:constraint_pair}) and (\ref{eq:constraint_clique_lp}), we can arrange the Lagrangian as follows:
		  \begin{align}\label{eq:lp_lagrangian_alt}
		  L({\boldsymbol{\alpha},\boldsymbol{\mu},\boldsymbol{\beta},\boldsymbol{\gamma}},\mathbf{y,w,v}) &= \sum_{a=1}^N\sum_{i \in \mathcal{L}}(\phi_{a:i}- \beta_a - \gamma_a){y_{a:i}} + \frac{1}{2\lambda}\sum_{a=1}^N\sum_{i \in \mathcal{L}}(y_{a:i} - y^k_{a:i})^2 + \sum_{a=1}^N\beta_a \\\nonumber& 
		  + \sum_{a=1}^N\sum_{\substack{b=1 \\ b \neq a}}^N\sum_{i \in \mathcal{L}}(\alpha^1_{ab:i} - \alpha^2_{ab:i} - \alpha^1_{ba:i} + \alpha^2_{ba:i})y_{a:i}\\\nonumber&
		  + \sum_{p=1}^R\sum_{\substack{c,d \in \mathcal{R}_p \\ c\neq d }}\sum_{i \in \mathcal{L}}(\mu_{cd:pi}^1 - \mu_{cd:pi}^2 - \mu_{dc:pi}^1 + \mu_{dc:pi}^2)y_{c:i}.
		  \end{align}
		  From this, differentiating the Lagrangian with respect to $\mathbf{y}$ is a trivially achieved:
		  \begin{align}
		  \frac{1}{\lambda}(y_{a:i} - y^k_{a:i}) =& - \sum_{\substack{b=1 \\ b \neq a}}^N\sum_{i \in \mathcal{L}}(\alpha^1_{ab:i} - \alpha^2_{ab:i} - \alpha^1_{ba:i} + \alpha^2_{ba:i}) + \beta_a + \gamma_{a:i} \\\nonumber &- \sum_{p=1}^R\sum_{\substack{d \in \mathcal{R}_p \\ a \neq d }}(\mu_{ad:pi}^1 - \mu_{ad:pi}^2 - \mu_{da:pi}^1 + \mu_{da:pi}^2)  - \phi_{a:i}.
		  \end{align}
		  By utilising the matrices introduced in equations (\ref{eq:deff_A}), (\ref{eq:deff_U}) and (\ref{eq:deff_B}) this expression can be concisely written in vector form:
		  \begin{align}\label{eq:y_proj}
		  \mathbf{y} = \lambda(\mathbf{A}\boldsymbol{\alpha} + \mathbf{U}\boldsymbol{\mu} + \mathbf{B}\boldsymbol{\beta} + \boldsymbol{\gamma} - \boldsymbol{\phi}) + \mathbf{y}^k.
		  \end{align}
		  Substituting this equation into the Lagrangian defined in (\ref{eq:lp_lagrangian_alt}) yields the following Lagrange dual problem:
		  \begin{align}
\min_{\boldsymbol{\alpha},\boldsymbol{\mu},\boldsymbol{\beta},\boldsymbol{\gamma}} g(\boldsymbol{\alpha},\boldsymbol{\mu},\boldsymbol{\beta},\boldsymbol{\gamma}) =& \frac{\lambda}{2}|| \mathbf{A}\boldsymbol{\alpha} + \mathbf{U}\boldsymbol{\mu} + \mathbf{B}\boldsymbol{\beta} + \boldsymbol{\gamma} - \boldsymbol{\phi}||^2 + \langle\mathbf{A}\boldsymbol{\alpha} + \mathbf{U}\boldsymbol{\mu} + \mathbf{B}\boldsymbol{\beta} + \boldsymbol{\gamma} - \boldsymbol{\phi}, \mathbf{y}^k \rangle \\\nonumber - &\langle\mathbf{1},\boldsymbol{\beta}\rangle\\
\text{s.t }\indent \gamma_{a:i}  \geq& \text{  } 0  \text{  } \forall a \in \{1,...,N\} \text{  } \forall i \in \mathcal{L}\nonumber,\\	
\boldsymbol{\alpha} \in \mathcal{A} =& \left\{
\boldsymbol{\alpha}\text{  }
\begin{tabular}{|cccc}\nonumber
$\alpha_{ab:i}^1 + \alpha_{ab:i}^2 = \frac{K_{ab}}{2}$ & $ a,b \in \{1,...,N\}, a \neq b,  i \in \mathcal{L}$ \\
$\alpha_{ab:i}^1,\alpha_{ab:i}^2 \geq 0$ & $  a,b \in \{1,...,N\}, a \neq b,  i \in \mathcal{L}$ \\
\end{tabular}
\right\},\\
\boldsymbol{\mu} \in \mathcal{U} =& \left\{
\boldsymbol{\mu}\text{  }\nonumber
\begin{tabular}{|c}
$\sum_{i \in \mathcal{L}}\sum_{\substack{c,d \in \mathcal{R}_p \\ c \neq d}}\mu_{cd:pi}^1 + \mu_{cd:pi}^2 = C_{p}\quad p \in \{1,...,R\}$\\
$\mu_{cd:pi}^1, \mu_{cd:pi}^2 \geq 0$  \quad   $ c,d \in \mathcal{R}_p, c \neq d, i \in \mathcal{L}, p \in \{1,...,R\} $\\
\end{tabular}
\right\}.\\ \nonumber
		  \end{align}
		  
		  \subsection{Closed form expression for $\boldsymbol{\beta}$}\label{app:beta_op}
		  Due to the unconstrained nature of $\boldsymbol{\beta}$, the minimum value of the dual objective $g$ is obtained when $\nabla_\beta g(\boldsymbol{\alpha^t},\boldsymbol{\mu^t},\boldsymbol{\beta},\boldsymbol{\gamma}) = 0$ and hence $\boldsymbol{\beta}$ can be derived as a function of $\boldsymbol{\gamma}$. Using the fact that $\mathbf{B}^T\mathbf{y}^k = \mathbf{1}, $ $ \nabla_{\boldsymbol{\beta}}g(\cdot)$ can be written as: 
		  \begin{align}
		  \nabla_{\boldsymbol{\beta}}g(\boldsymbol{\alpha}^t,\boldsymbol{\mu}^t,\boldsymbol{\beta},\boldsymbol{\gamma}) = \lambda\mathbf{B}^T(\mathbf{A}\boldsymbol{\alpha} + \mathbf{U}\boldsymbol{\mu} + \mathbf{B}\boldsymbol{\beta} + \boldsymbol{\gamma} - \boldsymbol{\phi})
		  \end{align}
		   Using $\mathbf{B}^T\mathbf{B} = M\mathbf{I}$ and the fact that $\lambda$ is a constant, an expression for $\boldsymbol{\beta}$ can be given as:
		   \begin{align}
		   \boldsymbol{\beta} = - \frac{\mathbf{B}^T}{M}(\mathbf{A}\boldsymbol{\alpha^t} + \mathbf{U}\boldsymbol{\mu^t} + \boldsymbol{\gamma} - \boldsymbol{\phi})
		   \end{align}
		  
		  \subsection{Quadratic program for $\boldsymbol{\gamma}_a$}\label{app:gamma_op}
		  By substituting the expression for $\boldsymbol{\beta}$ into the dual objective (\ref{eq:lp_dual}), a quadratic optimisation problem over $\boldsymbol{\gamma}$ is formed:
		  \begin{align}
		  	\min_{\boldsymbol{\gamma}} g(\boldsymbol{\alpha}^t,\boldsymbol{\mu}^t,\boldsymbol{\gamma}) &= \frac{\lambda}{2}|| \mathbf{D}(\mathbf{A}\boldsymbol{\alpha} + \mathbf{U}\boldsymbol{\mu} + \boldsymbol{\gamma} - \boldsymbol{\phi})||^2  + \langle\mathbf{D}(\mathbf{A}\boldsymbol{\alpha} + \mathbf{U}\boldsymbol{\mu} + \boldsymbol{\gamma} - \boldsymbol{\phi}), \mathbf{y}^k \rangle \\\nonumber & +  \frac{\mathbf{B}^T}{M}\langle\mathbf{1},\mathbf{A}\boldsymbol{\alpha^t} + \mathbf{U}\boldsymbol{\mu^t} + \boldsymbol{\gamma} - \boldsymbol{\phi}\rangle, 
		  \end{align}
		  where $\mathbf{D} = \boldsymbol{I} - \frac{\mathbf{BB}^T}{M}$. Using the fact that $\mathbf{B}^T\mathbf{y}^k = \mathbf{1}$, the identity $\mathbf{D}\mathbf{D}^T = \mathbf{D}$ and removing constant terms, the optimisation problem over $\boldsymbol{\gamma}$ can be simplified:
		  \begin{align}
		  \min_{\boldsymbol{\gamma} } \indent&\frac{\lambda}{2}\boldsymbol{\gamma^T}\mathbf{D}\boldsymbol{\gamma} + \langle \lambda\mathbf{D}((\mathbf{A}\boldsymbol{\alpha^t}) + (\mathbf{U}\boldsymbol{\mu^t}) - \boldsymbol{\phi}) + \mathbf{y}^k,\boldsymbol{\gamma}\rangle,\\
		  \text{s.t}\indent & \boldsymbol{\gamma} \geq \mathbf{0}.\nonumber
		  \end{align} 
		  Due to the fact that $\boldsymbol{D}$ is a block diagonal, the resulting problem can be written as a sum of quadratic programs:
		\begin{align}
		\min_{\boldsymbol{\gamma} \geq 0} g(\boldsymbol{\alpha}^t,\boldsymbol{\mu}^t,\boldsymbol{\gamma}) = &\sum_a\min_{\gamma \geq 0}\frac{1}{2}\boldsymbol{\gamma^T}_a\mathbf{Q}\boldsymbol{\gamma}_a + \langle \mathbf{Q}((\mathbf{A}\boldsymbol{\alpha^t})_a + (\mathbf{U}\boldsymbol{\mu^t})_a - \boldsymbol{\phi}_a) + \mathbf{y}^k_a,\boldsymbol{\gamma}_a\rangle,\\
		\text{s.t}\indent & \boldsymbol{\gamma}_a \geq \mathbf{0}.\nonumber
		\end{align} 
		Which can be optimised independently:
		\begin{align}
		\min_{\boldsymbol{\gamma_a}} \indent&\frac{1}{2}\boldsymbol{\gamma^T}_a\mathbf{Q}\boldsymbol{\gamma}_a + \langle \mathbf{Q}((\mathbf{A}\boldsymbol{\alpha^t}_a) + (\mathbf{U}\boldsymbol{\mu^t})_a - \boldsymbol{\phi}_a) + \mathbf{y}^k_a,\boldsymbol{\gamma}_a\rangle,\\
		\text{s.t}\indent & \boldsymbol{\gamma}_a \geq \mathbf{0}.\nonumber
		\end{align} 
			Here, $\boldsymbol{\gamma}_a$ denotes the vector $\{\gamma_{a:i}|i \in \mathcal{L}\}$ and $\mathbf{Q} = \lambda(\boldsymbol{I} - \frac{\mathbf{1}}{M}) \in \mathbb{R}^{M \times M}$. Thus the resulting optimisation problem decomposes to $N$ subproblems, with each subproblem being an $M$ dimensional QP.
			
		  \subsection{Derivation of the conditional gradient for $\mathbf{s_\alpha}$ and $\mathbf{s_\mu}$}\label{app:grad_comp_lp}
		  As previously stated, efficient conditional gradient computation is critical to a well performing Frank Wolfe algorithm. This appendix details the method for computing the conditional gradient in linear time. Attention is drawn to the computation of the conditional gradient $\mathbf{s_\alpha}$, in which a modified version of the filter-based method detailed in section \ref{subsec:permuto_lattice} is used.  A summary is provided in Appendix \ref{app:filter}.
		  \paragraph{Derivation of the conditional gradient for $\mathbf{s_\alpha}$}
		  With the dual variables fixed at $\boldsymbol{\alpha}^t,\boldsymbol{\mu}^t,\boldsymbol{\beta}^t,\boldsymbol{\gamma}^t$ the conditional gradient with respect to $\boldsymbol{\alpha}$ is obtained via solving the following:
		  \begin{align}
		  \mathbf{s}_\alpha = \argmin_{\mathbf{s}_\alpha \in\mathcal{A}}\langle\mathbf{s}_\alpha,\nabla_{\boldsymbol{\alpha}}g(\boldsymbol{\alpha}^t,\boldsymbol{\mu}^t,\boldsymbol{\beta}^t,\boldsymbol{\gamma}^t)\rangle.
		  \end{align} 
		  By using equation(\ref{eq:y_proj}), $\nabla_{\boldsymbol{\alpha}}g(\cdot)$ is given as:
		  \begin{align}
		  	\nabla_{\boldsymbol{\alpha}}g(\boldsymbol{\alpha}^t,\boldsymbol{\mu}^t,\boldsymbol{\beta}^t,\boldsymbol{\gamma}^t) = \mathbf{A}^T\tilde{\mathbf{y}}^t.
		  \end{align}
		  Note that the feasible set $\mathcal{A}$ is separable and can be written as $\mathcal{A} = \prod_{a,b \neq a,i}\mathcal{A}_{ab:i}$, where $\mathcal{A}_{ab:i} = \{(\alpha^1_{ab:i},\alpha^2_{ab:i}) | \alpha^1_{ab:i} + \alpha^2_{ab:i} = \frac{1}{2}K_{ab}, \alpha^1_{ad:i},\alpha^2_{ab:i} \geq 0\}$. It is possible exploit this separability and compute the conditional gradient $\mathbf{s_\alpha}$ for each Lagrange multiplier as follows:
		  \begin{align}
		  \min_{s_{\alpha^1_{ab:i}},s_{\alpha^2_{ab:i}}} & s_{\alpha^1_{ab:i}}\nabla_{\alpha^1_{ad:i}}g(\boldsymbol{\alpha}^t,\boldsymbol{\mu}^t,\boldsymbol{\beta}^t,\boldsymbol{\gamma}^t) + 
		  s_{\alpha^2_{ab:i}}\nabla_{\alpha^2_{ad:i}}g(\boldsymbol{\alpha}^t,\boldsymbol{\mu}^t,\boldsymbol{\beta}^t,\boldsymbol{\gamma}^t), \\
		  \text{s.t}\indent&  s_{\alpha^1_{ab:i}},s_{\alpha^2_{ab:i}} \in \mathcal{A}_{ab:i}.\nonumber
		  \end{align}
		  The derivatives $\nabla_{\alpha^1_{ad:i}}g(\cdot)$ and $\nabla_{\alpha^2_{ad:i}}g(\cdot)$ can be easily computed to yield the following:
		  \begin{align}
		  \nabla_{\alpha^1_{ab:i}}g(\boldsymbol{\alpha}^t,\boldsymbol{\mu}^t,\boldsymbol{\beta}^t,\boldsymbol{\gamma}^t) &= \tilde{y}^t_{b:i} - \tilde{y}^t_{a:i}\\
		  \nabla_{\alpha^2_{ab:i}}g(\boldsymbol{\alpha}^t,\boldsymbol{\mu}^t,\boldsymbol{\beta}^t,\boldsymbol{\gamma}^t) &= \tilde{y}^t_{a:i} - \tilde{y}^t_{b:i},
		  \end{align}
		  where the reader is reminded that $\tilde{y}^t_{a:i}$ represents the current infeasible solution, as detailed in step 6 of Algorithm \ref{algo:lp}. Hence, the minimum is given as:
		  \begin{align}
		  s_{\alpha^1_{ab:i}} = 
		  \begin{cases}
		  K_{ab}/2 & \text{if  } \tilde{y}^t_{a:i} \geq \tilde{y}^t_{b:i}\\
		  0 & \text{otherwise,}
		  \end{cases}\\
		  s_{\alpha^2_{ab:i}} = 
		  \begin{cases}
		  K_{ab}/2 & \text{if  } \tilde{y}^t_{a:i} \leq \tilde{y}^t_{b:i}\\
		  0 & \text{otherwise.}
		  \end{cases}
		  \end{align}
		  Which by utilising matrix $\mathbf{A}$, introduced in equation (\ref{eq:deff_A}), can be concisely written as:
		  \begin{align}
		  (\mathbf{As_\alpha})_{a:i} = \sum_{b}(K_{ab}\mathbbm{1}[\tilde{y}^t_{a:i} \leq \tilde{y}^t_{b:i}] - K_{ab}\mathbbm{1}[\tilde{y}^t_{b:i} \leq \tilde{y}^t_{a:i}]),
		  \end{align}
		  which can be solved efficiently using a modified version of the filter-based method. More details of this modified filter-based method are given in Appendix \ref{app:filter}.
		  
		  \paragraph{Derivation of the conditional gradient $\mathbf{s_\mu}$}
		  Similarly to $\mathbf{s_\mu}$ the conditional gradient of $\boldsymbol{\mu}$ at $\boldsymbol{\alpha}^t,\boldsymbol{\mu}^t,\boldsymbol{\beta}^t,\boldsymbol{\gamma}^t$ is obtained by via solving the following:
		  \begin{align}
		  \mathbf{s}_\mu = \argmin_{\mathbf{s}_\mu \in\mathcal{U}}\langle\mathbf{s}_\mu,\nabla_{\boldsymbol{\mu}}g(\boldsymbol{\alpha}^t,\boldsymbol{\mu}^t,\boldsymbol{\beta}^t,\boldsymbol{\gamma}^t)\rangle.
		  \end{align}  
			By using equation(\ref{eq:y_proj}), $\nabla_{\boldsymbol{\mu}}g(\cdot)$ is given as:
			\begin{align}
			\nabla_{\boldsymbol{\mu}}g(\boldsymbol{\alpha}^t,\boldsymbol{\mu}^t,\boldsymbol{\beta}^t,\boldsymbol{\gamma}^t) = \mathbf{U}^T\tilde{\mathbf{y}}^t.
			\end{align}
		  The set $\mathcal{U}$ can only be separated according to the number of cliques, $\mathcal{U} = \prod_p\mathcal{U}_p$, where $\mathcal{U}_{p} = \{(\mu^1_{cd:pi},\mu^2_{cd:pi}) | \sum_{i \in \mathcal{L}}\sum_{c,d}\mu_{cd:pi}^1 + \mu_{cd:pi}^2 = C_{p} ,\mu_{cd:pi}^1, \mu_{cd:pi}^2 \geq 0,  c,d \neq c,  i \in \mathcal{L}\}$. The conditional gradient for each set $\mathcal{U}_{p}$ can be written as:
		  \begin{align}\label{eq:decom_cond_u}
		  \min_{s_{\mu^1_{cd:pi}},s_{\mu^2_{cd:pi}}} & s_{\mu^1_{cd:pi}}\nabla_{\mu^1_{cd:pi}}g(\boldsymbol{\alpha}^t,\boldsymbol{\mu}^t,\boldsymbol{\beta}^t,\boldsymbol{\gamma}^t) + 
		  s_{\mu^2_{cd:pi}}\nabla_{\mu^2_{cd:pi}}g(\boldsymbol{\alpha}^t,\boldsymbol{\mu}^t,\boldsymbol{\beta}^t,\boldsymbol{\gamma}^t), \\
		  \text{s.t}\indent&  s_{\mu^1_{cd:pi}},s_{\mu^2_{cd:pi}} \in \mathcal{U}_{p}.\nonumber
		  \end{align}
		  The derivatives $\nabla_{\mu^1_{cd:pi}}g(\cdot)$ and $\nabla_{\mu^2_{cd:pi}}g(\cdot)$ can be easily computed to yield the following:
		  \begin{align}
		  \nabla_{\mu^1_{cd:pi}}g(\boldsymbol{\alpha}^t,\boldsymbol{\mu}^t,\boldsymbol{\beta}^t,\boldsymbol{\gamma}^t) &= \tilde{y}^t_{d:pi} - \tilde{y}^t_{c:pi},\\
		  \nabla_{\mu^2_{cd:pi}}g(\boldsymbol{\alpha}^t,\boldsymbol{\mu}^t,\boldsymbol{\beta}^t,\boldsymbol{\gamma}^t) &= \tilde{y}^t_{c:pi} - \tilde{y}^t_{d:pi},
		  \end{align}
		  where the reader is reminded that $\tilde{y}^t_{a:i}$ represents the current infeasible solution, as detailed in step 6 of Algorithm \ref{algo:lp}. Given $s_{\mu^1_{cd:pi}},s_{\mu^2_{cd:pi}} \in \mathcal{U}_{p}$, the conditional gradients are thus given as:
		  \begin{align}
		  s_{\mu^1_{cd:pi}} = 
		  \begin{cases}
		  C_{p}/2 & \text{if  } \tilde{y}^t_{c:pi} = \max_{c \in p, i \in \mathcal{L}}\tilde{y}^t_{c:pi} \text{ and } \tilde{y}^t_{d:i} = \min_{d \in p, i \in \mathcal{L}}\tilde{y}^t_{d:pi}\\
		  0 & \text{otherwise,}
		  \end{cases}\\
		  s_{\mu^2_{cd:pi}} = 
		  \begin{cases}
		  C_{p}/2 & \text{if  } \tilde{y}^t_{c:pi} = \min_{c \in p, i \in \mathcal{L}}\tilde{y}^t_{c:pi} \text{ and } \tilde{y}^t_{d:i} = \max_{d \in p, i \in \mathcal{L}}\tilde{y}^t_{d:pi}\\
		  0 & \text{otherwise.}
		  \end{cases}
		  \end{align}
		  By utilising the matrix $\mathbf{U}$ the conditional gradient of $\mathbf{s}_\mu$ can be written as:
		  \begin{align}
		  (\mathbf{Us_\mu})_{c:pi} = \begin{cases}
		  C_p & \text{if  } \tilde{y}^t_{c:pi} \leq \tilde{y}^t_{d:pj} \indent \forall d \in p \indent \forall j \in \mathcal{L}\\
		  -C_p & \text{if  } \tilde{y}^t_{c:pi} > \tilde{y}^t_{d:pj} \indent \forall d \in p \indent \forall j \in \mathcal{L}\\
		  0 & \text{otherwise}.
		  \end{cases}
		  \end{align}
		  
		  \subsection{Optimal step size}\label{app:lp_step}
		  We need to find the step size $\delta$ that gives the maximum
		  decrease in the objective function $g$ given the descent direction $\mathbf{s}^t$.  
		  % Once the step size is found, the variables $\bfalpha$ are updated as follows:
		  % \begin{align}
		  % \bfalpha^{t+1} &= \bfalpha^t + \delta\left(\bfs^t-\bfalpha^t\right)\ ,\\\nonumber
		  % A\bfalpha^{t+1} &= A\bfalpha^t + \delta\left(A\bfs^t-A\bfalpha^t\right)\ .
		  % \end{align}
		  This can be formalized as the following optimization problem:
		  \begin{align}
		  \underset{\delta} {\operatorname{min}}\quad &
		  \frac{\lambda}{2}\left\|\mathbf{A}\boldsymbol{\alpha}^t + \delta(\mathbf{As}^t_\alpha - \mathbf{A}\boldsymbol{\alpha}^t) + \mathbf{U}\boldsymbol{\mu}^t + \delta(\mathbf{Us}^t _\mu - \mathbf{U}\boldsymbol{\mu}^t) + 
		  \mathbf{B}\boldsymbol{\beta}^t + \boldsymbol{\gamma}^t - \boldsymbol{\phi}^t \right\|^2  \\\nonumber & + \left\langle \mathbf{A}\boldsymbol{\alpha}^t + \delta(\mathbf{As}^t_\alpha - \mathbf{A}\boldsymbol{\alpha}^t) + \mathbf{U}\boldsymbol{\mu}^t + \delta(\mathbf{Us}^t _\mu - \mathbf{U}\boldsymbol{\mu}^t) + 
		  \mathbf{B}\boldsymbol{\beta}^t + \boldsymbol{\gamma}^t - \boldsymbol{\phi}^t , \mathbf{y}^k
		  \right\rangle \\\nonumber & - \langle\mathbf{1},\boldsymbol{\beta}^t\rangle
		  ,\\\nonumber \text{s.t.}\quad \delta & \in[0,1] \ .
		  \end{align}
		  Note that the above function is optimized over the scalar variable $\delta$ and
		  the minimum is attained when the derivative is zero. Hence setting the derivate to zero:
		  \begin{align}
		  0 &=   \left\langle
		  \mathbf{y}^k, \mathbf{As}^t_\alpha - \mathbf{A}\boldsymbol{\alpha}^t + \mathbf{Us}^t _\mu - \mathbf{U}\boldsymbol{\mu}^t \right\rangle\ + \\\nonumber &  \lambda\left\langle (1 - \delta)\mathbf{A}\boldsymbol{\alpha}^t + \delta\mathbf{As}^t_\alpha + (1 - \delta)\mathbf{U}\boldsymbol{\mu}^t + \delta\mathbf{Us}^t _\mu + 
		  \mathbf{B}\boldsymbol{\beta}^t + \boldsymbol{\gamma}^t - \boldsymbol{\phi}^t, \mathbf{As}^t_\alpha - \mathbf{A}\boldsymbol{\alpha}^t + \mathbf{Us}^t _\mu - \mathbf{U}\boldsymbol{\mu}^t \right\rangle \\\nonumber  
		  \delta &= \frac{\langle \mathbf{As}^t_\alpha - \mathbf{A}\boldsymbol{\alpha}^t + \mathbf{Us}^t _\mu - \mathbf{U}\boldsymbol{\mu}^t,  \lambda\left(\mathbf{A}\boldsymbol{\alpha}^t + \mathbf{U}\boldsymbol{\mu}^t +
		  	\mathbf{B}\boldsymbol{\beta}^t + \boldsymbol{\gamma}^t - \boldsymbol{\phi}^t + \mathbf{y}^k \right\rangle}{\lambda\|\langle \mathbf{As}^t_\alpha - \mathbf{A}\boldsymbol{\alpha}^t + \mathbf{Us}^t _\mu - \mathbf{U}\boldsymbol{\mu}^t\|^2}\ ,\\\nonumber
		  \delta & = P_{[0,1]}\Bigg[\frac{\langle\mathbf{A}\boldsymbol{\alpha}^t + \mathbf{U}\boldsymbol{\mu}^t  - \mathbf{As}^t_\mu -  \mathbf{Us}^t _\alpha ,\mathbf{{y}^t}\rangle}{\lambda||\mathbf{A}\boldsymbol{\alpha}^t + \mathbf{U}\boldsymbol{\mu}^t  - \mathbf{As}^t_\mu -  \mathbf{Us}^t _\alpha||^2}\Bigg].
		  \end{align}
		  In fact, if the
		  optimal $\delta$ is out of the interval $[0,1]$ it is simply projected back.

		  \subsection{Filtering Method Appendix}\label{app:filter}

		  \paragraph{Original filtering algorithm}
		  Let us first introduce some notations below. We denote the set of lattice points
		  of the original permutohedral lattice with $\mathcal{P}$ and
		  the neighbouring feature points of lattice point $l$ by $N(l)$. Also we denote the
		  neighbouring lattice points of a feature point $a$ by $\overline{N}(a)$. In addition,
		  the barycentric weight between the lattice point $l$ and feature point $b$ is
		  denoted with $w_{lb}$. Furthermore, the value at feature point $b$ is denoted
		  with $v_b$ and the value at lattice point $l$ is denoted with $\overline{v}_{l}$. The
		  pseudocode of the algorithm is given in Algorithm~\ref{alg:phold}
		  % %%%% OLD-PH %%%%
		  \begin{algorithm}[H]
		  	\caption{Original filtering algorithm~\cite{FastHigh-DimensionalFilteringUsingthePermutohedralLattice}}
		  	\label{alg:phold}
		  	\begin{algorithmic}%[1]
		  		\Require Permutohedral lattice $\mathcal{P}$
		  		\ForAll{$l \in P$}
		  		\State{$\overline{v}_l \gets \sum_{b\in N(l)} w_{lb}\, v_b$}\Comment{Splatting}
		  		\EndFor
		  		\State $\overline{V}' \gets k \otimes \overline{V}$ \Comment{Blurring}
		  		
		  		\ForAll{$a \in \{1,...,N\}$}
		  		\State{$v'_a \gets \sum_{l\in \overline{N}(a)} w_{la}\,
		  			\overline{v}'_l$}\Comment{Slicing}
	  			\EndFor
		  		
		  	\end{algorithmic}
		  \end{algorithm}
		  
		  \paragraph{Modified filtering algorithm}\label{app:phnew}
		  As mentioned in the main paper the interval $[0,1]$ is discretized into $H$
		  discrete bins. Note that each bin $h\in \{0\ldots H\}$ is associated with a
		  probability interval which is identified as:
		  $\left[\frac{h}{H-1},\frac{h+1}{H-1}\right)$.
		  To this end, the bin $h_b$ of the feature point $b$ satisfy the following
		  inequality: $\left[\frac{h_b}{H-1} \le y_b < \frac{h_b+1}{H-1}\right]$.
		  
		  Furthermore, at the splatting step, the values $v_b$ are accumulated to its
		  neighbouring lattice point only if the lattice point is above or equal to the
		  feature point level.
		  Formally, the barycentric interpolation at lattice point $l$ at level $h$ can be written as
		  \begin{equation}
		  \label{eqn:splat}
		  \overline{v}_{l:h} = \sum_{\substack{b\in N(l)\\ h \ge h_b}} w_{lb}\,v_b  = \sum_{b\in N(l)}
		  w_{lb}\,v_b\mathbbm{1}\left[\frac{h}{H-1}\ge y_b\right]\ ,
		  \end{equation}
		  where $h_b$ is the level of feature point $b$ and $w_{lb}$ is the barycentric
		  weight between lattice point $l$ and feature point $b$. Then blurring is performed
		  independently at each discrete level $h$. Finally, at the slicing step, the
		  resulting values are interpolated at the level of the feature point.
		  Our modified algorithm is given in Algorithm~\ref{alg:phnew}.
		  
		  %%%% NEW-PH %%%%
		  \begin{algorithm}[H]
		  	\caption{Modified filtering algorithm}
		  	\label{alg:phnew}
		  	\begin{algorithmic}%[1]
		  		\Require Permutohedral lattice $\mathcal{P}$, discrete levels $H$
		  		\ForAll{$l \in \mathcal{P}$}\Comment{Splatting}
		  		\ForAll{$h \in \{0\ldots H-1\}$}
		  		\State $\overline{v}_{l:h} \gets \sum_{b\in N(l)}
		  		w_{lb}\,v_b\mathbbm{1}\left[\frac{h}{H-1}\ge y_b\right]$
		  		\EndFor
		  		\EndFor
		  		
		  		\ForAll{$h \in \{0\ldots H-1\}$} 
		  		\State{$\overline{V}'_h \gets k \otimes
		  			\overline{V}_h$}\Comment{Blurring}
	  			\EndFor
		  		
		  		\ForAll{$a \in \{1,...,N\}$}\Comment{Slicing}
		  		\State $v'_a \gets \sum_{l\in \overline{N}(a)} \sum_{h=0}^{H-1}
		  		w_{la}\,\overline{v}'_{l:h}\mathbbm{1}\left[\frac{h}{H-1} \le y_a < \frac{h+1}{H-1}\right]$
		  		\EndFor
		  		
		  	\end{algorithmic}
		  \end{algorithm}
		  Note that, the above algorithm is given for the $\mathbbm{1}[y_a\ge y_b]$ constraint, however it is fairly easy to modify it for the $\mathbbm{1}[y_a\le
		  y_b]$ constraint. In particular, one needs to change the interval identified by
		  the bin $h$ to: $\left(\frac{h-1}{H-1},\frac{h}{H-1}\right]$. Using this fact,
		  one can easily derive the splatting and slicing equations for the $\mathbbm{1}[y_a\le
		  y_b]$ constraint.
		  % one needs to use the $1$-based bin index
		  % instead of $0$-based, \ie, $h\in\{1\ldots H\}$. Then
		  
		  \subsection{Cross Validated Parameters}\label{app:params}	
		  A table of the cross validated parameters are given in Table \ref{tab:params}.
			  	\begin{table}\centering
			  		\begin{tabular}{ c @{ } | c c c c c c c c}
			  			\toprule
			  			Algorithm & $\sigma^{(3)}$ & $w^{(2)}$ & $\sigma^{(1)}$ & $\sigma^{(2)}$ & $w^{(1)}$ & $clique size$ & $\Gamma$ & $\eta$ \\\hline
			  			\multicolumn{9}{c}{MSRC} \\\hline
			  			\textbf{MF5} &  4.10 & 77.047 &  47.79 &  4.69 &  100 & - & -&  - \\
			  			\textbf{QP} & 2.36 &  22.89 &  48.73 &  6.52 & 60.50 & - & -& - \\
			  			\textbf{MF5\textsubscript{clique}} & 6.53 & 4.46 & 50 & 9.74 &  11.56 & 10 & 54.88 &    876.08 \\
			  			\textbf{QP\textsubscript{clique}} & 3.74 & 17.67 & 39.76 & 9.49 & 54.56 & 100 & 19.71 & 109.0 \\\bottomrule

			  			\multicolumn{9}{c}{Pascal} \\\hline
			  			\textbf{MF5} & 1.00 & 29.19 & 17.82 & 6.14 & 32.56 & - & -&  - \\
			  			\textbf{QP} & 1.00 & 100 & 19.11 & 6.08 & 55.19 & - & -& - \\
			  			\textbf{MF5\textsubscript{clique}} & 1.20 & 76.38 & 16.32 & 38.10 & 1.45 & 27 & 20.71 & 467.36  \\
			  			\textbf{QP\textsubscript{clique}} & 1.00 & 99.53 & 13.30 & 7.89 & 100.00 & 97 & 100 & 139.70 \\\bottomrule
			  		\end{tabular}\\\caption{Table of cross validated parameters for Pascal and MSRC.\label{tab:params}}
			  	\end{table}

		  \subsection{Additional Results}\label{app:res}	  
\begin{figure*}
				\def \SUBWIDTH{0.12\linewidth}
				\begin{center}	
					\begin{subfigure}{\SUBWIDTH}
						\includegraphics[width=0.99\linewidth]{Media/MSRC/image/4_3_s.png}
						%					\caption{Image}
					\end{subfigure}%
					\begin{subfigure}{\SUBWIDTH}
						\includegraphics[width=0.99\linewidth]{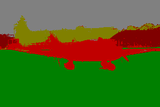}
						%				\caption{MF5}
					\end{subfigure}%
					\begin{subfigure}{\SUBWIDTH}
						\includegraphics[width=0.99\linewidth]{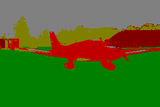}
						%			\caption{QP}
					\end{subfigure}%
					\begin{subfigure}{\SUBWIDTH}
						\includegraphics[width=0.99\linewidth]{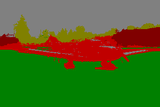}
						%					\caption{LP}
					\end{subfigure}%
					\begin{subfigure}{\SUBWIDTH}
						\includegraphics[width=0.99\linewidth]{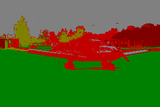}
						%					\caption{MF5}
					\end{subfigure}%
					\begin{subfigure}{\SUBWIDTH}
						\includegraphics[width=0.99\linewidth]{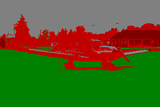}
						%					\caption{QP}
					\end{subfigure}%
					\begin{subfigure}{\SUBWIDTH}
						\includegraphics[width=0.99\linewidth]{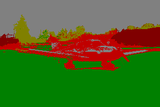}
						%					\caption{LP}
					\end{subfigure}%
					\begin{subfigure}{\SUBWIDTH}
						\includegraphics[width=0.99\linewidth]{Media/MSRC/gt/4_3_s_GT.png}
						%					\caption{GT}
					\end{subfigure}
					
					\begin{subfigure}{\SUBWIDTH}
						\includegraphics[width=0.99\linewidth]{Media/MSRC/image/17_15_s.png}
						%					\caption{Image}
					\end{subfigure}%
					\begin{subfigure}{\SUBWIDTH}
						\includegraphics[width=0.99\linewidth]{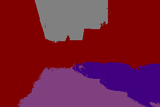}
						%				\caption{MF5}
					\end{subfigure}%
					\begin{subfigure}{\SUBWIDTH}
						\includegraphics[width=0.99\linewidth]{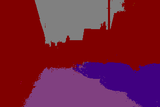}
						%			\caption{QP}
					\end{subfigure}%
					\begin{subfigure}{\SUBWIDTH}
						\includegraphics[width=0.99\linewidth]{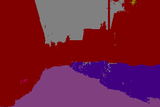}
						%					\caption{LP}
					\end{subfigure}%
					\begin{subfigure}{\SUBWIDTH}
						\includegraphics[width=0.99\linewidth]{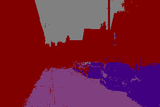}
						%					\caption{MF5}
					\end{subfigure}%
					\begin{subfigure}{\SUBWIDTH}
						\includegraphics[width=0.99\linewidth]{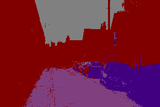}
						%					\caption{QP}
					\end{subfigure}%
					\begin{subfigure}{\SUBWIDTH}
						\includegraphics[width=0.99\linewidth]{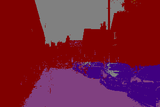}
						%					\caption{LP}
					\end{subfigure}%
					\begin{subfigure}{\SUBWIDTH}
						\includegraphics[width=0.99\linewidth]{Media/MSRC/gt/17_15_s_GT.png}
						%					\caption{GT}
					\end{subfigure}
					
					\begin{subfigure}{\SUBWIDTH}
						\includegraphics[width=0.99\linewidth]{Media/MSRC/image/5_23_s.png}
						%					\caption{Image}
					\end{subfigure}%
					\begin{subfigure}{\SUBWIDTH}
						\includegraphics[width=0.99\linewidth]{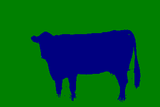}
						%				\caption{MF5}
					\end{subfigure}%
					\begin{subfigure}{\SUBWIDTH}
						\includegraphics[width=0.99\linewidth]{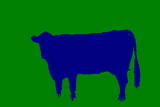}
						%			\caption{QP}
					\end{subfigure}%
					\begin{subfigure}{\SUBWIDTH}
						\includegraphics[width=0.99\linewidth]{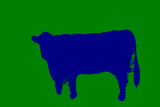}
						%					\caption{LP}
					\end{subfigure}%
					\begin{subfigure}{\SUBWIDTH}
						\includegraphics[width=0.99\linewidth]{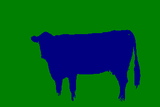}
						%					\caption{MF5}
					\end{subfigure}%
					\begin{subfigure}{\SUBWIDTH}
						\includegraphics[width=0.99\linewidth]{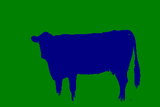}
						%					\caption{QP}
					\end{subfigure}%
					\begin{subfigure}{\SUBWIDTH}
						\includegraphics[width=0.99\linewidth]{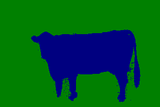}
						%					\caption{LP}
					\end{subfigure}%
					\begin{subfigure}{\SUBWIDTH}
						\includegraphics[width=0.99\linewidth]{Media/MSRC/gt/5_23_s_GT.png}
						%					\caption{GT}
					\end{subfigure}

					%% pascal
					
										\begin{subfigure}{\SUBWIDTH}
						\includegraphics[width=0.99\linewidth]{Media/Pascal/image/2007_005331.jpg}
						%					\caption{Image}
					\end{subfigure}%
					\begin{subfigure}{\SUBWIDTH}
						\includegraphics[width=0.99\linewidth]{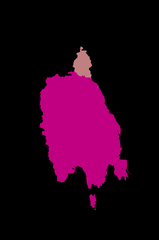}
						%				\caption{MF5}
					\end{subfigure}%
					\begin{subfigure}{\SUBWIDTH}
						\includegraphics[width=0.99\linewidth]{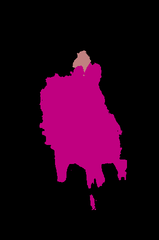}
						%			\caption{QP}
					\end{subfigure}%
					\begin{subfigure}{\SUBWIDTH}
						\includegraphics[width=0.99\linewidth]{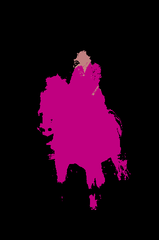}
						%					\caption{LP}
					\end{subfigure}%
					\begin{subfigure}{\SUBWIDTH}
						\includegraphics[width=0.99\linewidth]{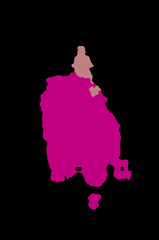}
						%					\caption{MF5}
					\end{subfigure}%
					\begin{subfigure}{\SUBWIDTH}
						\includegraphics[width=0.99\linewidth]{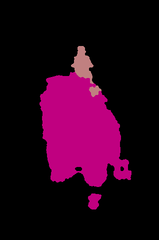}
						%					\caption{QP}
					\end{subfigure}%
					\begin{subfigure}{\SUBWIDTH}
						\includegraphics[width=0.99\linewidth]{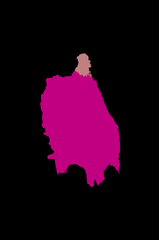}
						%					\caption{LP}
					\end{subfigure}%
					\begin{subfigure}{\SUBWIDTH}
						\includegraphics[width=0.99\linewidth]{Media/Pascal/gt/2007_005331.png}
						%					\caption{GT}
					\end{subfigure}
				
					\begin{subfigure}{\SUBWIDTH}
						\includegraphics[width=0.99\linewidth]{Media/Pascal/image/2009_003005.jpg}
	%					\caption{Image}
					\end{subfigure}%
					\begin{subfigure}{\SUBWIDTH}
						\includegraphics[width=0.99\linewidth]{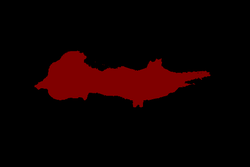}
		%				\caption{MF5}
					\end{subfigure}%
					\begin{subfigure}{\SUBWIDTH}
						\includegraphics[width=0.99\linewidth]{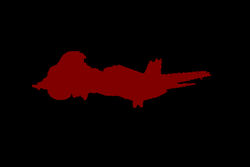}
			%			\caption{QP}
					\end{subfigure}%
					\begin{subfigure}{\SUBWIDTH}
						\includegraphics[width=0.99\linewidth]{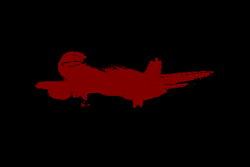}
	%					\caption{LP}
					\end{subfigure}%
					\begin{subfigure}{\SUBWIDTH}
						\includegraphics[width=0.99\linewidth]{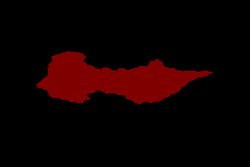}
	%					\caption{MF5}
					\end{subfigure}%
					\begin{subfigure}{\SUBWIDTH}
						\includegraphics[width=0.99\linewidth]{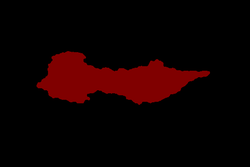}
	%					\caption{QP}
					\end{subfigure}%
					\begin{subfigure}{\SUBWIDTH}
						\includegraphics[width=0.99\linewidth]{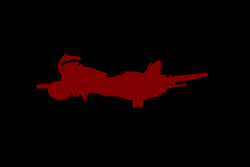}
	%					\caption{LP}
					\end{subfigure}%
					\begin{subfigure}{\SUBWIDTH}
						\includegraphics[width=0.99\linewidth]{Media/Pascal/gt/2009_003005.png}
	%					\caption{GT}
					\end{subfigure}

					\begin{subfigure}{\SUBWIDTH}
						\includegraphics[width=0.99\linewidth]{Media/Pascal/image/2007_000676.jpg}
						\caption*{Image}
					\end{subfigure}%
					\begin{subfigure}{\SUBWIDTH}
						\includegraphics[width=0.99\linewidth]{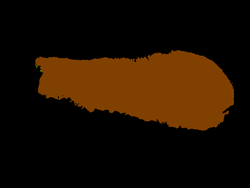}
						\caption*{MF5}
					\end{subfigure}%
					\begin{subfigure}{\SUBWIDTH}
						\includegraphics[width=0.99\linewidth]{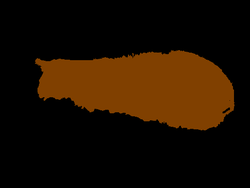}
						\caption*{QP}
					\end{subfigure}%
					\begin{subfigure}{\SUBWIDTH}
						\includegraphics[width=0.99\linewidth]{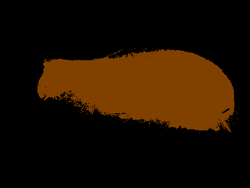}
						\caption*{LP}
					\end{subfigure}%
					\begin{subfigure}{\SUBWIDTH}
						\includegraphics[width=0.99\linewidth]{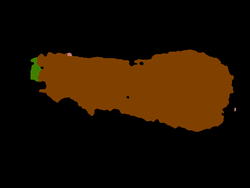}
						\caption*{MF5\textsubscript{clique}}
					\end{subfigure}%
					\begin{subfigure}{\SUBWIDTH}
						\includegraphics[width=0.99\linewidth]{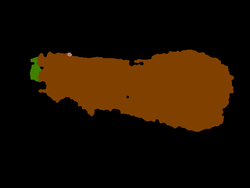}
						\caption*{QP\textsubscript{clique}}
					\end{subfigure}%
					\begin{subfigure}{\SUBWIDTH}
						\includegraphics[width=0.99\linewidth]{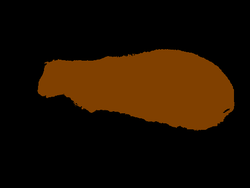}
						\caption*{LP\textsubscript{clique}}
					\end{subfigure}%
					\begin{subfigure}{\SUBWIDTH}
						\includegraphics[width=0.99\linewidth]{Media/Pascal/gt/2007_000676.png}
						\caption*{GT}
					\end{subfigure}
					
					\vspace{-0.2cm}
					\caption{\em Qualitative results with the parameters
						tuned for \textbf{MF5\textsubscript{clique}} and \textbf{MF5}. As can be clearly seen, even though the parameters have been tuned to \textbf{MF5}, the \textbf{LP} algorithm produces competitive segmentations when compared to \textbf{MF5}.}\label{app:mf_images}
				\end{center}
				\vspace{-0.9cm}
			\end{figure*}

		  In this section we present the results for when the algorithms are tuned to \textbf{MF5\textsubscript{clique}} and \textbf{MF}, displayed in Table \ref{tab:mf5_energies}. As is consistent with our previous results, it can be seen that \textbf{QP} achieves lower energies then \textbf{MF5}, but fails to reach the low energies of \textbf{LP}. A similar pattern can be seen for the higher-order potentials - where \textbf{LP\textsubscript{clique}} obtains lower energies then \textbf{QP\textsubscript{clique}}. Qualitative results can be seen in Figure \ref{app:mf_images}.

		  	 %%% accuracy table %%%
		  	 \begin{table}[h]\centering
		  	 	\begin{tabular}{@{}ccccc@{}}
		  	 		\toprule
		  	 		Algorithm                         & \textbf{Avg.E ($\times10^7$)} & \textbf{Time(s)} & \textbf{Acc(\%)} & \textbf{IoU(\%)} \\\hline
		  	 		\multicolumn{5}{c}{Pascal} \\\hline
		  	 		\textbf{MF5}                      & 5.26  & \textbf{0.35} & 79.54 & \textbf{22.23} \\
		  	 		\textbf{QP}                       & 4.13  & 1.06 & {79.63} & 22.23\\
		  	 		\textbf{LP}                       & \textbf{1.17}  & 54.0 & \textbf{79.84} & 21.91\\\cline{1-2}
		  	 		\textbf{MF5\textsubscript{clique}} & 5.20  & 1.75 & 79.26 & {22.22}\\	 	
		  	 		\textbf{QP\textsubscript{clique}} & 3.78  & 1.75 & 79.25 & 22.21\\	 		
		  	 		\textbf{LP\textsubscript{clique}} & \textbf{2.19}  & 79.67 & 79.67 & {21.35}\\
		  	 		
		  	 		\toprule
		  	 		\multicolumn{5}{c}{MSRC} \\\hline
		  	 		\textbf{MF5}                      & 17.7  & \textbf{0.37} & 83.79 & 57.16 \\
		  	 		\textbf{QP}                       & 12.2  & 0.58 & {83.93} & 57.80\\
		  	 		\textbf{LP}                       & \textbf{0.39}  & 67.9 & 82.93 & 57.30\\\cline{1-2}
		  	 		\textbf{MF5\textsubscript{clique}} & 0.34  & 0.62 & 84.30 & \textbf{60.53}\\	
		  	 		\textbf{QP\textsubscript{clique}} & 0.30  & 1.13 & \textbf{84.41} & {60.32}\\	 		
		  	 		\textbf{LP\textsubscript{clique}} & \textbf{0.12}  & 46.9 & 81.89 & 55.66\\
		  	 		\bottomrule
		  	 	\end{tabular}\caption{\textit{Table displaying the average energy, timings, accuracy and IoU, when the parameters are tuned to \textbf{MF5\textsubscript{clique}} and \textbf{MF5}. It is shown that the lowest energies are achieved by \textbf{LP\textsubscript{clique}} and \textbf{LP}.Interestingly \textbf{LP} obtain the greatest segmentation accuracy for Pascal, despite tuning the parameters for \textbf{MF5\textsubscript{clique}} and \textbf{MF5}.}\label{tab:mf5_energies}}
		  	 \end{table}
		  	 
			%%% segmentation images %%%

		  \bibliographystyle{abbrv}
		  \bibliography{densecrf_ho}

		\end{document}